\documentclass[acmtog]{acmart}
\acmSubmissionID{153}

\usepackage{booktabs} 

\usepackage{bbding}
\usepackage{color}
\usepackage[ruled]{algorithm2e} 

\SetAlFnt{\small}
\SetAlCapFnt{\small}
\SetAlCapNameFnt{\small}
\SetAlCapHSkip{0pt}

\acmJournal{TOG}

\begin{document}
\title{DeepVecFont: Synthesizing High-quality Vector Fonts via Dual-modality Learning}
\author{Yizhi Wang}
\affiliation{%
  \institution{Wangxuan Institute of Computer Technology, Peking University}
  \country{China}}
\email{wangyizhi@pku.edu.cn}

\author{Zhouhui Lian}
\authornote{Corresponding author}
\affiliation{%
  \institution{Wangxuan Institute of Computer Technology, Peking University}
  \country{China}}
\email{lianzhouhui@pku.edu.cn}

\begin{abstract}

Automatic font generation based on deep learning has aroused a lot of interest in the last decade. However, only a few recently-reported approaches are capable of directly generating vector glyphs and their results are still far from satisfactory.
In this paper, we propose a novel method, DeepVecFont, to effectively resolve this problem. Using our method, for the first time, visually-pleasing vector glyphs whose quality and compactness are both comparable to human-designed ones can be automatically generated.
The key idea of our DeepVecFont is to adopt the techniques of image synthesis, sequence modeling and differentiable rasterization to exhaustively exploit the dual-modality information (i.e., raster images and vector outlines) of vector fonts.
The highlights of this paper are threefold. First, we design a dual-modality learning strategy which utilizes both image-aspect and sequence-aspect features of fonts to synthesize vector glyphs. 
Second, we provide a new generative paradigm to handle unstructured data (e.g., vector glyphs) by randomly sampling plausible synthesis results to get the optimal one which is further refined under the guidance of generated structured data (e.g., glyph images).
Finally, qualitative and quantitative experiments conducted on a publicly-available dataset demonstrate that our method obtains high-quality synthesis results in the applications of vector font generation and interpolation, significantly outperforming the state of the art.
\end{abstract}

%
%


%
%


\begin{teaserfigure}
  \includegraphics[width=\textwidth]{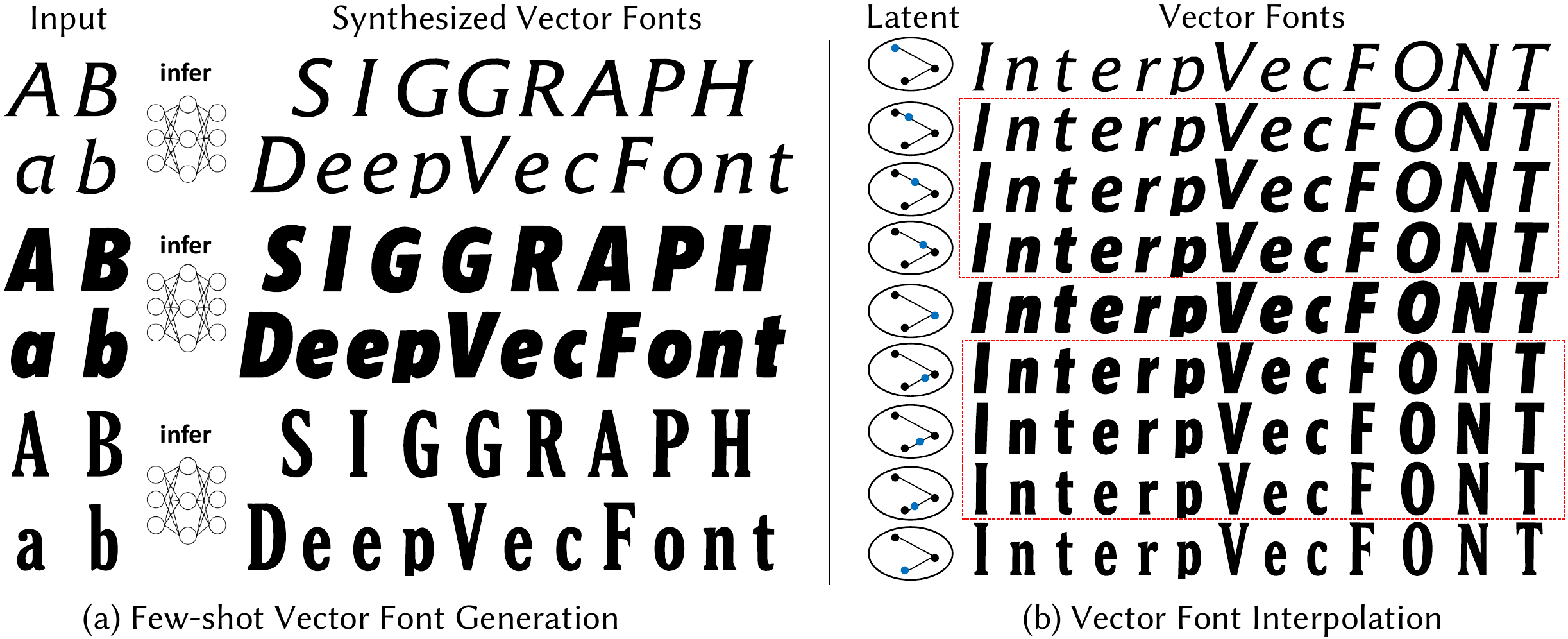}
  \caption{Different from most existing font generation methods that aim at synthesizing glyph images, our proposed DeepVecFont is capable of automatically generating high-quality vector fonts. (a) Given a few reference glyphs as input, our method can directly synthesize the whole vector font in the same style. (b) With the fonts synthesized in (a), more vector fonts can be generated by smooth interpolations in the style latent space of our model. The blue points indicate the locations of font style features in our latent space. The interpolated vector fonts are marked by red rectangles.}
  \Description{.}
  \label{fig:teaser}
\end{teaserfigure}

\maketitle

\section{Introduction}
Font generation is an important and challenging task in areas of Computer Graphics, Computer Vision and Artificial Intelligence. Despite the success of deep generative models in synthesizing glyph images~\cite{tian2017rewrite,jiang2017dcfont,azadi2018multi,gao2019artistic}, how to automatically generate high-quality vector fonts is now still an unsolved problem.
A traditional font designing procedure is to first obtain all characters’ raster images and then automatically/manually convert them to vector glyphs, which typically involves many pre-defined rules and still requires numerous manual interventions.
Early machine-learning based methods, such as ~\cite{suveeranont2010example,campbell2014learning}, perform affine transformations or manifold learning on existing font data to synthesize vector fonts in new styles. More recently, several methods based on sequential generative models have been proposed to directly generate the draw-command sequences of sketches, handwritings or SVGs (Scalable Vector Graphics), such as~\cite{ha2017neural,zhang2017drawing,tang2019fontrnn,lopes2019learned,carlier2020deepsvg}.
The sequential generative models are naturally suitable for vector font generation as they are capable of imitating the writing/drawing process of glyphs. 
However, there exist two critical problems in those previous methods that prevent them from synthesizing visually-pleasing vector glyphs.
\par
The first problem is that most of those previous works only consider one modality of glyphs when encoding features, i.e., using either the sequential representation (~\cite{ha2017neural,carlier2020deepsvg,zhang2017drawing,tang2019fontrnn}) or the image representation (~\cite{lopes2019learned}).
In this paper, we investigate how to fully exploit the information contained in the two modalities of glyphs, i.e., the image modality and the sequence modality, respectively. Given a vector glyph, the image modality provides us a global perception of the glyph's shape, while the sequence modality (i.e., drawing commands) offers us the concrete structural information and scale-invariant representation.
Furthermore, the overall structured information of a raster glyph image can be reconstructed well but its local details are often missing. On the contrary, a vector outline is typically composed of unstructured representations of many basic visual components (i.e., curves and lines) that contain rich and precise information for local regions.
In addition, machine learning (ML) models deal with the data of these two modalities in completely different manners, and thus can extract and learn complementary information from them. 
For vector glyphs, learning some font attributes (e.g., height, width, stroke width, etc.) usually needs RNNs to capture long-range dependencies in the time steps of draw-command sequence, which becomes more and more challenging when the length of range increases. However, in rasterized images, these attributes can be easily learnt by CNNs due to the utilization of convolutions with various receptive fields.
Therefore,
addressing the above-mentioned problem by designing a method to exhaustively exploit the dual-modality data is theoretically meaningful.
In our method, we first implement dual-modality feature fusion to provide richer information from both global and local views to accurately generate the drawing commands of glyphs.
Then, we also perform bidirectional translations between these two modalities to further exploit the information they contain.\par

The second problem is the location shift issue brought by Mixture Distribution Models.
Before discussing this issue, we give an explanation of why Mixture Distribution Models are necessary for handling this task, which has not been fully discussed in the previous work.
\begin{figure}[t!]
  \centering
  \includegraphics[width=\columnwidth]{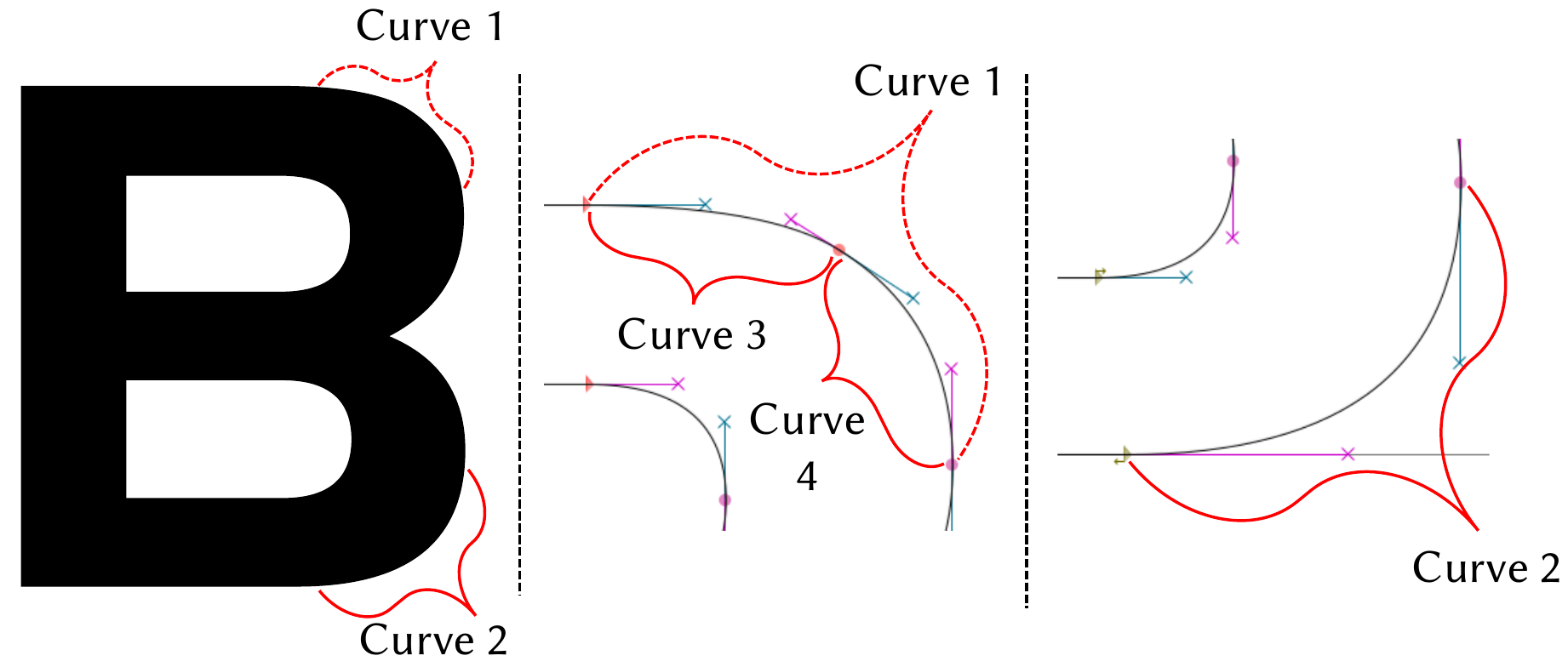}
  \caption{An explanation of why Mixture Distribution Models are needed for the training of drawing commands of glyphs. The Curve 1 and 2 in the glyph `B' are supposed to be identical, while in the vector font file they are differently represented. MDN attempts to eliminate the ambiguity by modeling several potential distributions of command coordinates.}
  \label{fig:WhyMDNNeeded}
\end{figure}
Unlike NLP (Natural Language Processing) tasks where each word in a sentence is discrete, the coordinates of drawing commands or trajectories are continuous and multi-valued (uncertain).
Take Fig.~\ref{fig:WhyMDNNeeded} as an example, the Curve 1 and 2 are supposed to be identical in view of the rendered glyph image. However, in the human-designed vector font file, the Curve 1 is composed of two cubic Bézier curves (the Curve 3 and 4) while the Curve 2 is represented using only one cubic Bézier curve.
We find that, in the vector glyph synthesizing task, simply minimizing the MSE (Mean Squared Error) and Cross Entropy (CE) losses is unable to precisely predict the coordinate values of control points on the contour, since the average of correct target values is often not a correct value (e.g., Fig.~\ref{fig:WhyMDNNeeded}). Thus, parameterizing a mixture distribution is a much better choice than minimizing those two losses.
The Mixture Density Network (MDN) was originally proposed in~\cite{bishop1994mixture} to handle the prediction of continuous variables in which the mapping to be learned is multi-valued.
Graves et al.~\shortcite{graves2013generating} then successfully applied MDN to Recurrent Neural Networks for the generation of handwriting sequences.
\par
\begin{figure}[t!]
  \centering
  \includegraphics[width=\columnwidth]{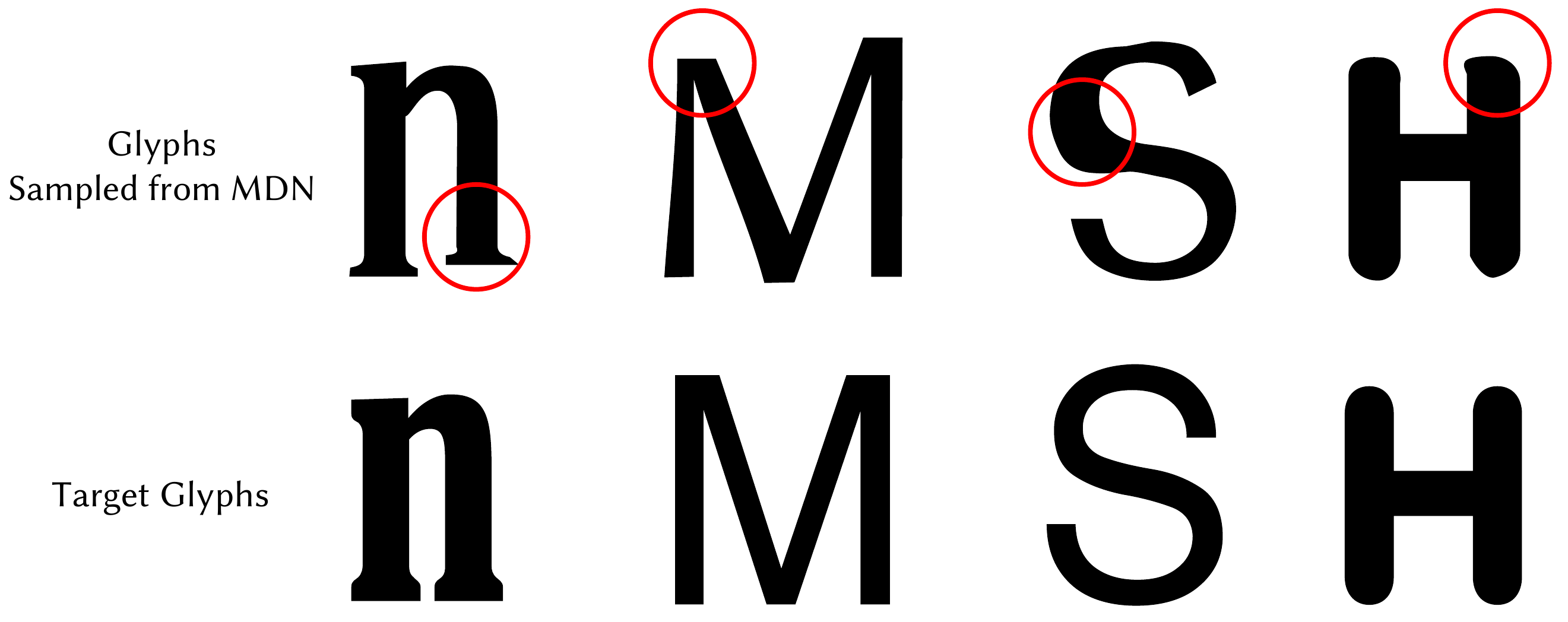}
  \caption{The problem of location shift brought by Mixture Density Networks. The first row demonstrates the glyphs generated by MDN in our model without refinement. The second row shows the target glyphs.}
  \label{fig:LocationShift}
\end{figure}

Although MDN is capable of coping with such kinds of uncertainties when training models, it tends to bring location shifts (see Fig.~\ref{fig:LocationShift}) to the outlines of glyphs because of the ``soft'' restriction (i.e., the mixture of several distributions).
Besides, RNNs suffer from the gradient vanishing problem and are struggling to build connections between different drawing steps.
Consequently, the generated vector glyphs cannot be strictly aligned with the ground truth, which is different against the image synthesis task.
To address the above-mentioned problem, we propose to employ differentiable rasterizers for imposing an additional restriction on the drawing commands predicted by MDN.
Specifically, it should be able to render the sequence of these drawing commands back to the target glyph image by differentiable rasterizers.
In the training phase, we design a neural approximate rasterizer, which is simple but effective, to impose this restriction.
In the inference stage, we adopt another more precise rasterizer~\cite{li2020differentiable} based on Monte Carlo sampling for further refinement to obtain our final synthesized vector glyphs.

\par

To sum up, major contributions of our paper are threefold.
\begin{enumerate}\setlength{\itemsep}{0pt} 

\item We design a dual-modality learning strategy which utilizes both image-aspect and sequence-aspect features of fonts to synthesize vector glyphs. Our experiments show that the joint exploitation of both modalities is necessary for the generation of high-quality vector fonts.
\item We provide a new generative paradigm to handle unstructured data (e.g., vector glyphs) by randomly sampling plausible synthesis results to get the optimal one which is further refined under the guidance of generated structured data (e.g., glyph images). We verify the effectiveness of the new paradigm in handling the challenging task of vector font synthesis.
\item We propose, DeepVecFont~\footnote{Source code is available at \url{https://github.com/yizhiwang96/deepvecfont}}, a novel vector font generation method based on the above two techniques. Using our method, for the first time, visually-pleasing vector glyphs whose quality and compactness are both comparable to human-designed ones can be automatically generated. Experiments demonstrate that our DeepVecFont obtains significantly better synthesis results compared to the state of the art.
\end{enumerate}


\section{Related Work}
\subsection{Glyph Image Synthesis}
GANs (Generative Adversarial Nets)~\cite{goodfellow2014generative} have emerged as one of the most popular neural generative models and have been widely applied in glyph image synthesis.
“zi2zi”~\cite{tian2017rewrite} aims to transfer a given glyph image in one font style to another based on a conditional GAN modified from ``pix2pix''~\cite{isola2017image}.
Jiang et al.~\shortcite{jiang2017dcfont} proposed an end-to-end font style transfer system to generate large-scale Chinese font libraries with only a small number of training samples.
Azadi et al.~\shortcite{azadi2018multi} proposed a stacked conditional GAN model to generate a set of multi-content glyph images following a consistent style from a few input samples.
Guo et al.~\shortcite{guo2018creating} managed to synthesize Chinese glyph images in new font styles by building font manifolds learned from existing font libraries.
Gao et al.~\shortcite{gao2019artistic} introduced a triple-discriminator GAN architecture (i.e., shape, texture and local discriminators) to further improve the local details of synthesized artistic glyph images.
Wang et al.~\shortcite{wang2020attribute2font} proposed a deep generative model to automatically create fonts by synthesizing glyph images according to user-specified attributes and their corresponding values.
Although all of these existing methods are capable of generating visually-pleasing glyph images, they fail to directly synthesize vector fonts. To get a high-quality font product, large amounts of manual interventions are still required to convert those glyph images into high-quality vector glyphs.
\subsection{Sequence Modeling}
Recurrent Neural Networks (RNNs) such as LSTM~\cite{hochreiter1997long} and GRU~\cite{cho2014learning} have been very successful in the NLP domain.
More recently, Transformers~\cite{vaswani2017attention} based on non-local feature representation become popular and outperform RNNs when trained with huge amounts of data.
Vector font synthesis can be viewed as a sequence generation problem where a drawing command is generated at each time step.
\subsection{Vector Font Generation}
Suveeranont and Igarashi~\shortcite{suveeranont2010example} presented a system to automatically generate the glyphs for all characters in the font library based on a glyph designed by the user. Their key idea is to automatically manipulate the outlines and skeletons of characters in a template font library, which is selected to have the most similar font style as the input glyph sample, to construct the glyphs for all characters in the desired font style.
Campbell and Kautz~\shortcite{campbell2014learning} proposed to synthesize fonts in unseen styles by learning a font mainfold based on the GP-LVM algrithm.
Lian et al.~\shortcite{lian2018easyfont} proposed a system to automatically generate large-scale Chinese handwriting fonts by learning styles of stroke shape and layout separately.
\par
Zhang et al.~\shortcite{zhang2017drawing} and Ha et al.~\shortcite{ha2017neural} utilized RNNs to generate handwritten characters or sketches.
Tang et al.~\shortcite{tang2019fontrnn} proposed to handle the task of generating large-scale Chinese fonts from a few input samples by using a modified RNN model with the monotonic attention mechanism.
Lopes et al.~\shortcite{lopes2019learned} developed SVG-VAE that employs an image autoencoder to learn font style features and then uses a RNN to generate vector glyphs.
Carlier et al.~\shortcite{carlier2020deepsvg} proposed DeepSVG which consists of a SVG autoencoder based on the hierarchical representations of SVGs and a Transformer-based encoder-decoder model.
Smirnov et al.~\shortcite{smirnov2020deep} utilized CNNs and distance fields to translate input glyph images into Bézier curves according to predetermined letter templates.
More recently, Reddy et al.~\shortcite{reddy2021im2vec} proposed a neural network model named Im2Vec that can generate vector graphics from raster training images without direct supervision of vector counterparts.\par
Here we give a detailed discussion about the differences among existing vector glyph synthesizing approaches including Font-MF~\cite{campbell2014learning}, SVG-VAE~\cite{lopes2019learned}, DeepSVG~\cite{carlier2020deepsvg}, Im2Vec~\cite{reddy2021im2vec}, and our method.
Table~\ref{tab:ModelComparison} shows the data modalities and formats in these methods, from which we can draw some conclusions as follows:
\begin{enumerate}\setlength{\itemsep}{0pt} 

\item Font-MF aims to generate new fonts by smooth interpolations between existing fonts, while our system can also take a few glyphs in a new font as input and synthesize the vector outlines of other remaining glyphs.

\item All existing methods simply consider one modality of glyph data in the encoding phase, i.e., using either the vector representation or the image representation.
On the one hand, encoding only the image modality inevitably brings more reconstruction errors due to the information loss brought by image rasterization.
On the other hand, encoding only the vector modality makes it tough for deep learning models (such as RNNs and Transformers) to learn a global perception for glyph shapes.
The dual-modality feature encoding we utilize is proven to be effective for high-quality font generation through extensive experiments.

\item For keypoint-based methods (Font-MF and Im2Vec), they need to first sample a number of points for each closed curve. Font-MF uses a fixed number (512) while Im2Vec predicts the number adaptively.
Font-MF needs to further precisely establish a dense correspondence between two glyphs. There exist some limitations in this type of approaches. For example, Font-MF cannot handle glyphs containing different numbers of closed curves and it fails to generate satisfactory results when the styles of some glyphs are relatively special. Moreover, Font-MF can only first generate glyphs consisting of polylines and then use a curve-fitting algorithm to convert them back to Bézier curves. Im2Vec employs only curves to reconstruct the shapes so that it cannot well fit the lines of glyphs. On the contrary, our system can be trained using the original vector fonts and directly synthesizes visually-pleasing vector outlines consisting of both lines and Bézier curves.

\end{enumerate}

\begin{table}[!htbp]
\centering
\caption{The comparison among recently proposed vector font (graphics) generation methods based on machine learning techniques. ``Img'' and ``Seq'' denote image and sequence, respectively. ``Seq Format'' denotes the sequence format. ``DR'' denotes differentiable rasterization. \Checkmark denotes the utilization of DR.
}
\label{tab:ModelComparison}
\scalebox{1.00}{
\begin{tabular}{|p{1.3cm}|p{1.5cm}|p{1.5cm}|p{1.5cm}|p{0.5cm}|}
\hline
  Method     & Encoding Modality     & Decoding Modality &  Seq Format &  DR \\
 \hline
 Font-MF   & Seq  &  Seq & Keypoints & - \\
 \hline
 SVG-VAE   & Img  & Img \& Seq & Commands & - \\
 \hline
 DeepSVG & Seq  & Seq & Commands &- \\
 \hline
 Im2Vec & Img  & Seq & Keypoints & \Checkmark \\
 \hline
 Ours & Img \& Seq   & Img \& Seq &Commands & \Checkmark \\
\hline
\end{tabular}
}

\end{table}

\par
\subsection{Multi-modality Representation Learning}
As mentioned in~\cite{baltruvsaitis2018multimodal}, multimodal representations can be roughly classified into the following two categories: joint and coordinated. Joint representations combine the unimodal signals into the same representation space, while coordinated representations process unimodal signals separately, but enforce certain similarity constraints on them to bring them to a coordinated space.
In our work, we leverage a joint representation of the image modality and sequence modality of glyphs.
After the encoding stage, image features and sequence features are first fused and then fed into the image decoder and the sequence decoder.
Representative methods for multimodal feature fusion include Concat+MLP, MCB~\cite{fukui2016multimodal}, MLB~\cite{kim2016hadamard} and MFB~\cite{yu2017multi}.
Concat+MLP (Concatenation + Multi-Layer Perceptron), one of the most intuitive strategies, is adopted in our model, which is proven to be effective through our experiments.
\subsection{Differentiable Rasterization}
Neural approximations of rasterization try to use neural networks to imitate the rasterization operation and make it differentiable~\cite{huang2019learning}.
More recently, Li et al.~\shortcite{li2020differentiable} developed a differentiable rasterizer which is enabled by two kinds of anti-aliasing approaches.
The first approach uses Monte Carlo sampling while the second one adopts an approximate analytical prefilter based on the signed distance to the the closest curve.
They also demonstrated how their method can further refine a SVG by minimizing the L1 loss between its rasterized image and the corresponding ground-truth image.
However, their method relies heavily on the initial vectorization results, and the sequence length and order of Bézier curves used in their method must be fixed. Therefore, it can not be directly used to generate high-quality vector glyphs which might contain arbitrary types and numbers of lines and curves. 
To address this problem, with neural generative models, we first obtain vector glyphs that consist of reasonable drawing commands in terms of the command length and categories.
Then we further refine the coordinates of each command with the help of the differentiable rasterizer proposed in~\cite{li2020differentiable}.

\section{Method description}

\subsection{Data structure}
\begin{figure}[t!]
  \centering
  \includegraphics[width=\columnwidth]{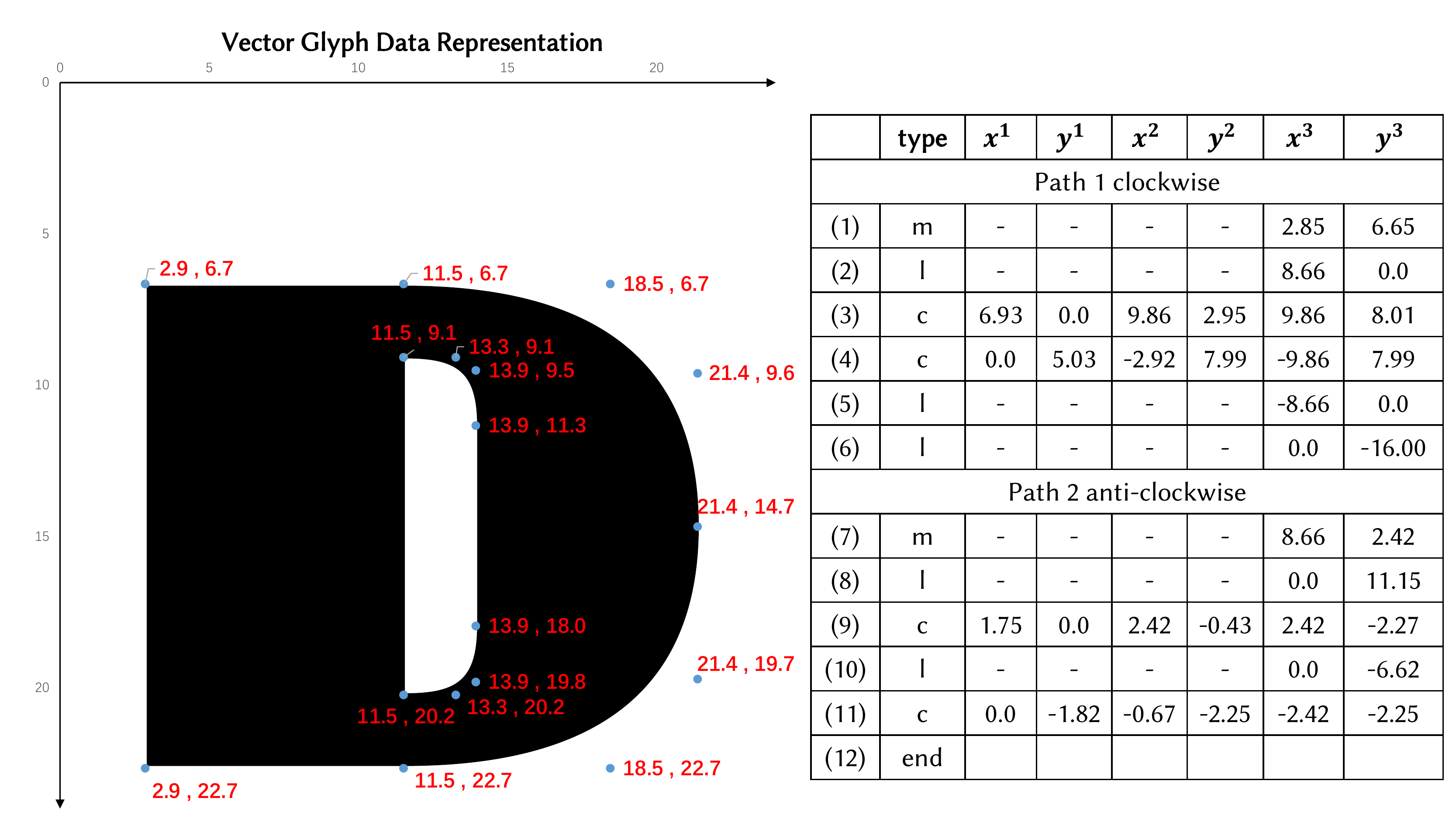}
  \caption{An illustration of our data structure. The right table shows the drawing commands of a glyph `D' (using relative coordinates). The left figure demonstrates the glyph shape and the argument points of drawing commands (using absolute coordinates). `m', `l' and `c' denote the $move$, $line$ and $curve$ commands, respectively. `-' denotes the unused argument. The clockwise or anti-clockwise order determines how to fill the outline.}
  \label{fig:DataStructure}
\end{figure}
The glyph set of a vector font can be represented as $\{G_{1}, ..., G_{N_{char}}\}$, where $N_{char}$ is the number of character categories. Typically, $N_{char} = 52$ for a glyph set consisting of `a'-`z' and `A'-`Z'.
$G_{i}$ denotes the drawing commands of the $i$-th glyph, which can be formulated as:
\begin{equation}
 G_{i}= \{C_{i,j} | 1 \leq j \leq l_{i}\},
\end{equation}
\begin{equation}
 C_{i,j}= (z_{i,j},p_{i,j}),
\end{equation}
\begin{equation}
p_{i,j}= \{(x_{i,j,k},y_{i,j,k}) | 1 \leq k \leq N_{p}\},
\end{equation}
where $l_{i}$ denotes the number of drawing commands in $G_{i}$, $C_{i,j}$ denotes the $j$-th command ($1 \leq j \leq l_{i}$) in $G(i)$, $z_{i,j}$ denotes the command type of $C_{i,j}$, $p_{i,j}$ means the coordinate arguments of $C_{i,j}$, and $N_{p}$ is the number of coordinate pairs in every command. We use a fixed-length argument list, where $N_{p}$ is fixed and any unused argument is set to 0. 
An illustration of our data structure is shown in Fig~\ref{fig:DataStructure}.\par
In this paper, we only consider 4 kinds of commands, i.e., $z_{i,j} \in \{move, line, curve, end\}$: (1) moving the drawing location (for starting a new path); (2) drawing a line; (3) drawing a three-order (cubic) Bézier curve; (4) ending the draw-command sequence.
Since the relative coordinate is employed and a cubic Bézier curve contains four control points, we set $N_{p}=3$ for all kinds of drawing commands. More specifically,
for $z_{i,j} \in \{move,line\}$, only $(x_{i,j,3},y_{i,j,3})$ is used which represents the ending (destination) point.
For $z_{i,j} = curve$, $(x_{i,j,1},y_{i,j,1})$ and $(x_{i,j,2},y_{i,j,2})$ denote the two control points and $(x_{i,j,3},y_{i,j,3})$ is the ending point.
For $z_{i,j} = end$, no argument is used.
For any unused argument, we set its value to 0.
When preparing data, we render the drawing commands to glyph images $\{X_{1}, ..., X_{N_{char}}\}$, i.e., ${X_{i}}$ is rasterized from ${C_{i}}$, by using traditional CG-based rasterization methods.\par
\subsection{Overview}
\begin{figure}[t!]
  \centering
  \includegraphics[width=\columnwidth]{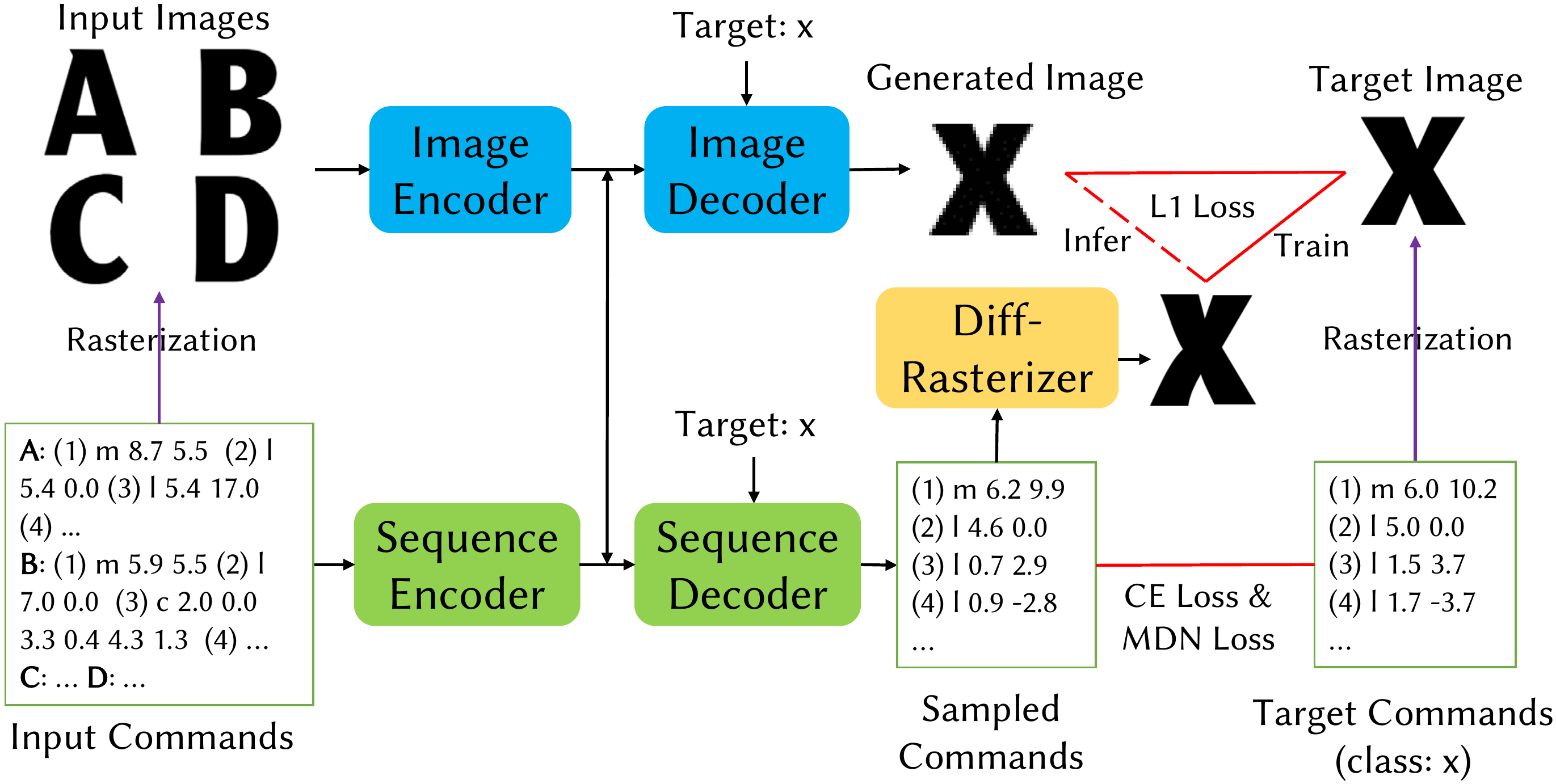}
  \caption{An overview of our proposed model. Given a few glyphs from a font, our model receives their drawing commands as input, and outputs the drawing commands of other glyphs in the same font style. We use relative coordinates for each command, and `m', `l' and `c' denote the $move$, $line$ and $curve$ commands, respectively. 
  ``Diff-Rasterizer'' denotes the Differentiable Rasterizer.
  In the training and inference stages, outputs of the Differentiable Rasterizer are refined by the target images and generated images respectively using the L1 loss.}
  \label{fig:Pipeline}
\end{figure}
We first briefly describe the pipeline of our method. As shown in Fig.~\ref{fig:Pipeline},
our model receives $N_{r}$ reference glyphs as input ($N_{r} = 4$ here) and outputs a target glyph.
The input glyphs and the ground truth of target glyphs are in the same font style, whose drawing commands are $\{C_{s_{i}}|1 \leq i \leq N_{r}\}$ and $C_{t}$, respectively, where $s_{i}$ and $t$ are randomly sampled from $\{1, 2, ...,  N_{char}\}$ in the training phase.
We first send the rendered images of input glyphs into the Image Encoder to learn the image-aspect style feature of their belonging font:
\begin{equation}
f^{img}= F_{enc}^{img}(X_{s_{1}}, X_{s_{2}}, ..., X_{s_{N_{r}}}).
\end{equation}
We also send the drawing commands of reference glyphs into the Sequence Encoder to learn the sequence-aspect style feature of their belonging font:
\begin{equation}
f^{seq}= F_{enc}^{seq}(G_{s_{1}}, G_{s_{2}}, ..., G_{s_{N_{r}}}).
\end{equation}
Then, we learn a unified representation $f$ for both of the image and sequence spaces by combining $f^{img}$ and $f^{seq}$.
Afterwards, we send the learnt feature $f$ into the Image Decoder to reconstruct the target glyph image:
\begin{equation}
\hat{X}_{t}= F_{dec}^{img}(f,t).
\end{equation}
Finally, we also send $f$ into the Sequence Decoder.
Since we utilize mixture distributions for command prediction, the output of the sequence decoder is not determinate.
Specifically, it outputs the potential distributions of the arguments of our expected glyph $\hat{G}_{t}$:
\begin{equation}
\Gamma_{t}= F_{dec}^{seq}(f,t),
\end{equation}
\begin{equation}
\hat{G}_{t}= (\hat{z}_{t},\hat{p}_{t}) \sim \Gamma_{t},
\end{equation}
i.e., the arguments of $\hat{G}_{t}$ are subjected to $\Gamma_{t}$.
\par
In both of the training and inference phases, we utilize differentiable rasterization to further refine the sampled glyphs to make them look more coordinated. Details will be revealed in the following sections.

\subsection{Image Encoder and Sequence Encoder}

\begin{figure}[t!]
  \centering
  \includegraphics[width=\columnwidth]{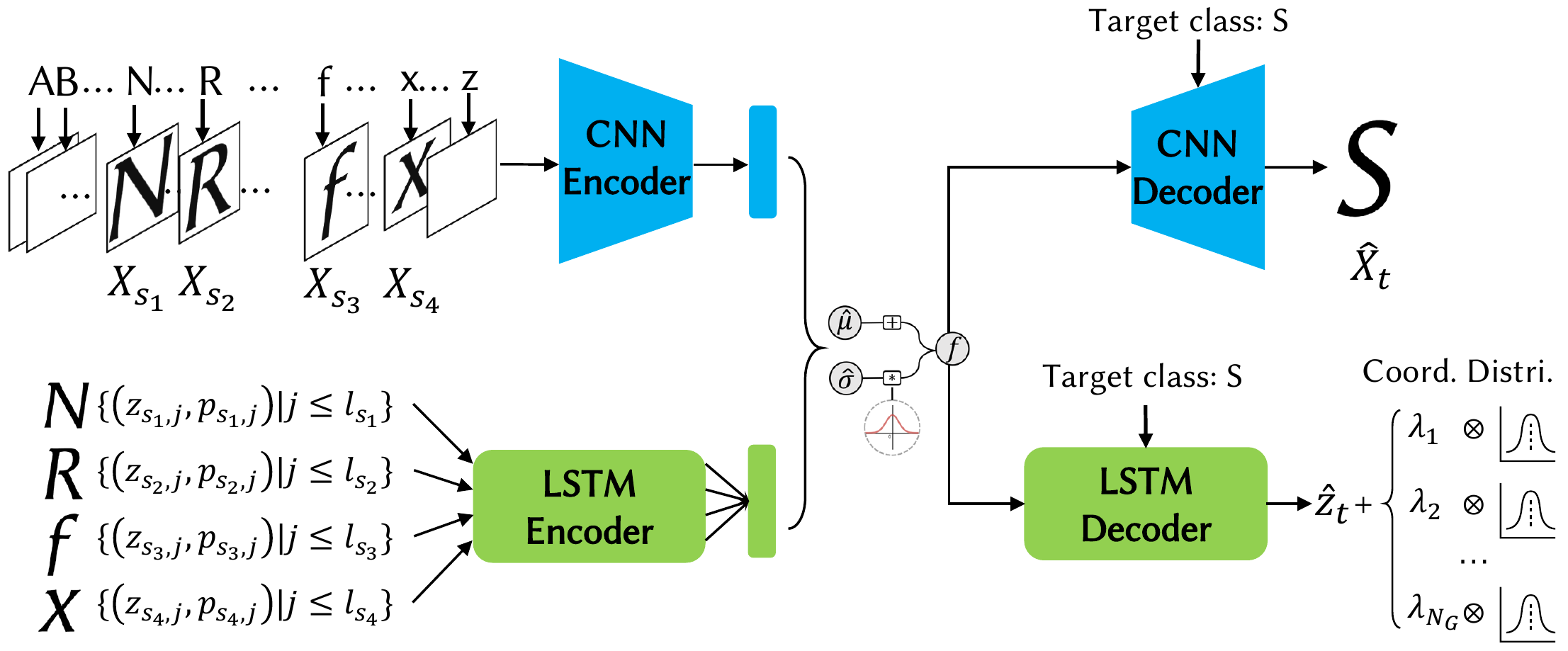}
  \caption{The detailed data flows of our Image Encoder-Decoder and Sequence Encoder-Decoder models.}
  \label{fig:ImageSeqEncDec}
\end{figure}
Typically, a single glyph is unable to sufficiently represent the style of its belonging font.
Thereby, we use $N$ reference glyphs to stabilize the statistics of font features extracted by the Image Encoder and the Sequence Encoder.
The Image Encoder is a Convolutional Neural Network (CNN).
Due to the parallel computation of each feature channel in CNNs, we concatenate all reference images in the channel dimension:
\begin{equation}
f^{img}= CNN([X_{s_{1}};X_{s_{2}};...;X_{s_{N_{r}}}]),
\end{equation}
where the square bracket denotes ordered concatenation, in other words, all these glyph images are concatenated according to their character categories and then fed into the Image Encoder (see Fig.~\ref{fig:ImageSeqEncDec}).

Previous methods only consider one modality of fonts when decoding font style features.
Although SVG-VAE~\cite{lopes2019learned} tries to translate the glyph image into its corresponding drawing commands, the sequence features of vector glyphs remain unexploited.
We notice that both sequence-aspect features and image-aspect features are beneficial for reconstructing vector glyphs.
In our model, we adopt LSTMs to build the Sequence Encoder and Decoder.
We encode the input sequences as follows.
First, we send each reference glyph into the Sequence Encoder separately and get its encoded feature:
\begin{equation}
f^{seq}_{s_{i}}= LSTMEnc(G_{s_{i}}).
\end{equation}
Next, the holistic sequence-aspect style feature is calculated by:
\begin{equation}
f^{seq} = W_{a}[f^{seq}_{s_{1}};f^{seq}_{s_{2}};...;f^{seq}_{s_{N_{r}}}],
\end{equation}
where $W_{a}$ is a linear projection which aggregates all the sequence features and the square bracket denotes concatenation.

\subsection{The Unified Latent Space}
We utilize a popular feature fusion strategy, i.e., Concat + MLP, to learn the joint representation of image and sequence modalities:
\begin{equation}
f = MLP([f^{img};f^{seq}]),
\end{equation}
where MLP denotes Multilayer Perceptron and the square bracket denotes concatenation.
The latent space is also normalized by the KL Loss introduced in VAE~\cite{kingma2013auto}:
\begin{equation}
L_{kl} = KL(f_{\mu},f_{\sigma},\mathcal{N}(0,I)),
\end{equation}
where $f_{\mu}$ and $f_{\sigma}$ are the predicted mean and standard deviation of $f$, respectively.
$L_{kl}$ attempts to make each dimension of $f$ follow the Gaussian Distribution $\mathcal{N}(0,1)$.
Benefiting from this, sampling in the latent space becomes easier and thus more new fonts can be generated.
\subsection{Image Decoder and Sequence Decoder}
With the learnt latent code $f$, we move forward to synthesize the target glyph images and vector fonts.
The Image Decoder is a Deconvolutional Neural Network (DCNN).
We send $f$ and the target character label into the Image Decoder:
\begin{equation}
\hat{X}_{t} = DCNN([f,t]).
\end{equation}
\par
Different from SVG-VAE, we employ the L1 loss instead of the Cross-Entropy loss for image reconstruction because the L1 loss results in much sharper images.
Besides, we also adopt the perceptual loss~\cite{johnson2016perceptual} as an addition term for calculating the image reconstruction loss $L_{rec}$:
\begin{equation}
L_{rec}= ||\hat{X}_{t} - {X}_{t}||_{1} + L_{percep}(\hat{X}_{t},{X}_{t}).
\end{equation}
The perceptual loss is defined by the feature distance in different layers of the VGG Network~\cite{simonyan2014very}, which is beneficial for preserving more details in the generated images.

The Sequence Decoder (i.e., an LSTM Decoder) receives the learnt feature and outputs the hidden state $h_{t}$ for the final prediction:
\begin{equation}
h_{t} = LSTMDec(f,t).
\end{equation}
A softmax classifier and a MDN model are attached to the last layer of the LSTM Decoder, to predict the command type and coordinates, respectively:
\begin{equation}
\hat{z}_{t} = Softmax(h_{t}),
\end{equation}
\begin{equation}
\{(\lambda_{k},\mu_{k}, \sigma_{k}) | 1\leq k \leq N_{G}\}= MDN(h_{t}),
\end{equation}
where $\mathcal{N}(\mu_{1}, \sigma_{N_{G}}^{2}),...,\mathcal{N}(\mu_{G}, \sigma_{N_{G}}^{2})$ are potential Gaussian Distributions for the coordinate values with normalized weights $\lambda_{1},...,\lambda_{N_{G}}$, i.e., $\lambda_{1} + ... + \lambda_{N_{G}}=1$.
Note that we predict $N_{G}$ distributions for every coordinate of each drawing command, and the corresponding subscripts are not listed here for brevity.
At last, there are two loss functions defined in this section:
\begin{equation}
L_{CE} = CE(\hat{z}_{t}, z_{t}),
\end{equation}
\begin{equation}
L_{MDN}= MDNLoss(p_{t}, MDN(h_{t})),
\end{equation}
where $CE$ denotes the Cross-Entropy loss and $MDNLoss$ measures how $p_{t}$ is deviated from these Gaussian Distributions.
The detailed calculation method of $L_{MDN}$ can be found in~\cite{bishop1994mixture}, which is skipped here since it is not the main contribution of this paper.

\subsection{Neural Differentiable Rasterizer}
The introduction of MDN inevitably brings location shifts and makes the generated vector glyphs look uncoordinated.
This is because the optimization of the Cross-Entropy and MDN losses lacks a global view of the glyph structure.
We want to leverage rasterizers to make direct alignment between the glyph image rendered from the predicted drawing commands and the corresponding ground truth during training.
There are two main reasons why~\cite{li2020differentiable}’s method cannot be integrated into our model during training:
(1) Their method requires that SVGs in a batch share the same numbers and lengths of paths, which is different from the setting of our model.
(2) In the early training stage, many predicted sequences of our model might be incomplete (there are absent or redundant ``move'' or ``end'' commands, which we call invalid sequences).
However,~\cite{li2020differentiable}’s method requires complete paths to perform rasterization.
Instead, we design an approximate Neural Differentiable Rasterizer (NDR) (denoted as $F_{ras}^{n}$) that is robust to the inputs to address this issue.
\subsubsection{Pre-training}
\begin{figure}[t!]
  \centering
  \includegraphics[width=\columnwidth]{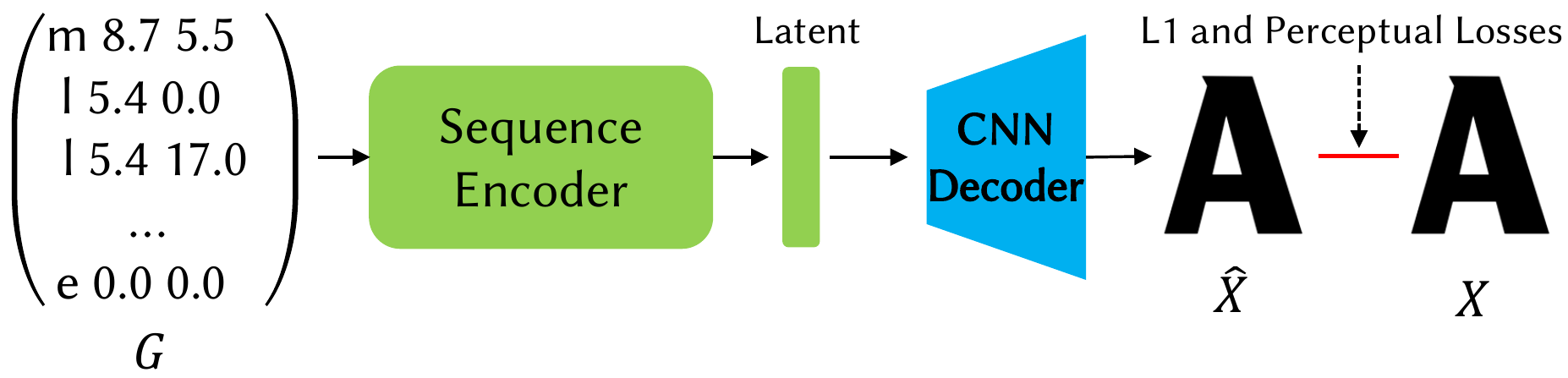}
  \caption{The architecture of our Neural Differentiable Rasterizer, which takes the draw-command sequence as input and outputs its corresponding image.}
  \label{fig:NeuralRasterizer}
\end{figure}

We train our Neural Differentiable Rasterizer, whose architecture is shown in Fig.~\ref{fig:NeuralRasterizer}, in advance of the above-mentioned main model.
The neural rasterizer is made up of another Sequence Encoder $F^{seq}_{enc'}$ and another Image Decoder $F^{img}_{dec'}$, which are different from those in the main model.
Given an arbitrary draw-command sequence $G$ in the training dataset, the rasterizer $F_{ras}^{n}$ learns to render its corresponding glyph image $\hat{X}$:
\begin{equation}
\hat{X} = F_{ras}^{n}(G) = F^{img}_{dec'}(F^{seq}_{enc'}(G)).
\end{equation}
The loss function is composed of the L1 Loss and the perceptual Loss.
Although rasterization results obtained by our neural rasterizer are not as precise as those generated by Monte-Carlo sampling methods, it still can output reasonable results when receiving invalid draw-command sequences sampled during the training phase. On the contrary, other methods like~\cite{li2020differentiable} will inevitably throw out errors in this situation.
Generally, the command-type classifier and MDN focus on the local while the neural rasterizer pays its attention to the global.
For this reason, the neural rasterizer provides meaningful supervision information in addition to the MDN and Cross-Entropy Losses.
The loss function for pre-training our neural rasterizer is defined as:
\begin{equation}
L_{ras}^{pre} = ||F_{ras}^{n}(G) -  X||_{1} + L_{p}(F_{ras}^{n}(G),X),
\end{equation}
where $L_{p}$ is the abbreviation of $L_{percep}$.
When the pre-training is finished, we plug the neural rasterizer into our main model to enhance the synthesis results of sequence prediction. 
\subsubsection{Integrated into the Main Model}
The parameters of our neural differentiable rasterizer are fixed when training our main model.
We randomly sample one set of drawing commands from our predicted distributions:
\begin{equation}
\hat{G}_{t} \sim GMM(\{(\lambda_{k},\mu_{k}, \sigma_{k}) | 1\leq k \leq N_{G}\}),
\end{equation}
where GMM denotes the Gaussian Mixture Model.
Although GMM is adopted to predict the command coordinates $\hat{p}_{t}$ in $\hat{G}_{t}$,  the command types $\hat{z}_{t}$ are also influenced by GMM.
This is because in sequence models the draw-command of a certain step is affected by previous movements.
Therefore, we adopt this formulation for brevity, i.e., GMM controls the sampling of both command types and coordinates.
The sampled sequence is then sent into the rasterizer $F_{ras}^{n}$.
We employ $L_{ras}$ as an important auxiliary loss to correctly reconstruct the global shapes of generated vector glyphs: 
\begin{equation}
L_{ras} = ||F_{ras}^{n}(\hat{G}_{t}) -  X_{t}||_{1}.
\end{equation}
Specifically, $L_{ras}$ is designed to stabilize the numerous Gaussian Distributions and avoid mode collapses in the training phase.
\subsection{The Loss Functions}

The whole loss function of our main model is formulated as:
\begin{equation}
L_{total} = L_{rec} + L_{CE} + L_{MDN} + L_{ras} + L_{kl}.
\end{equation}
The first term and the second to fourth terms are related to image reconstruction and sequence reconstruction, respectively. Here, the weight of each term is experimentally chosen and it is omitted in the formula for the sake of brevity. 
\subsection{Further Refinement}
\begin{figure}[t!]
  \centering
  \includegraphics[width=\columnwidth]{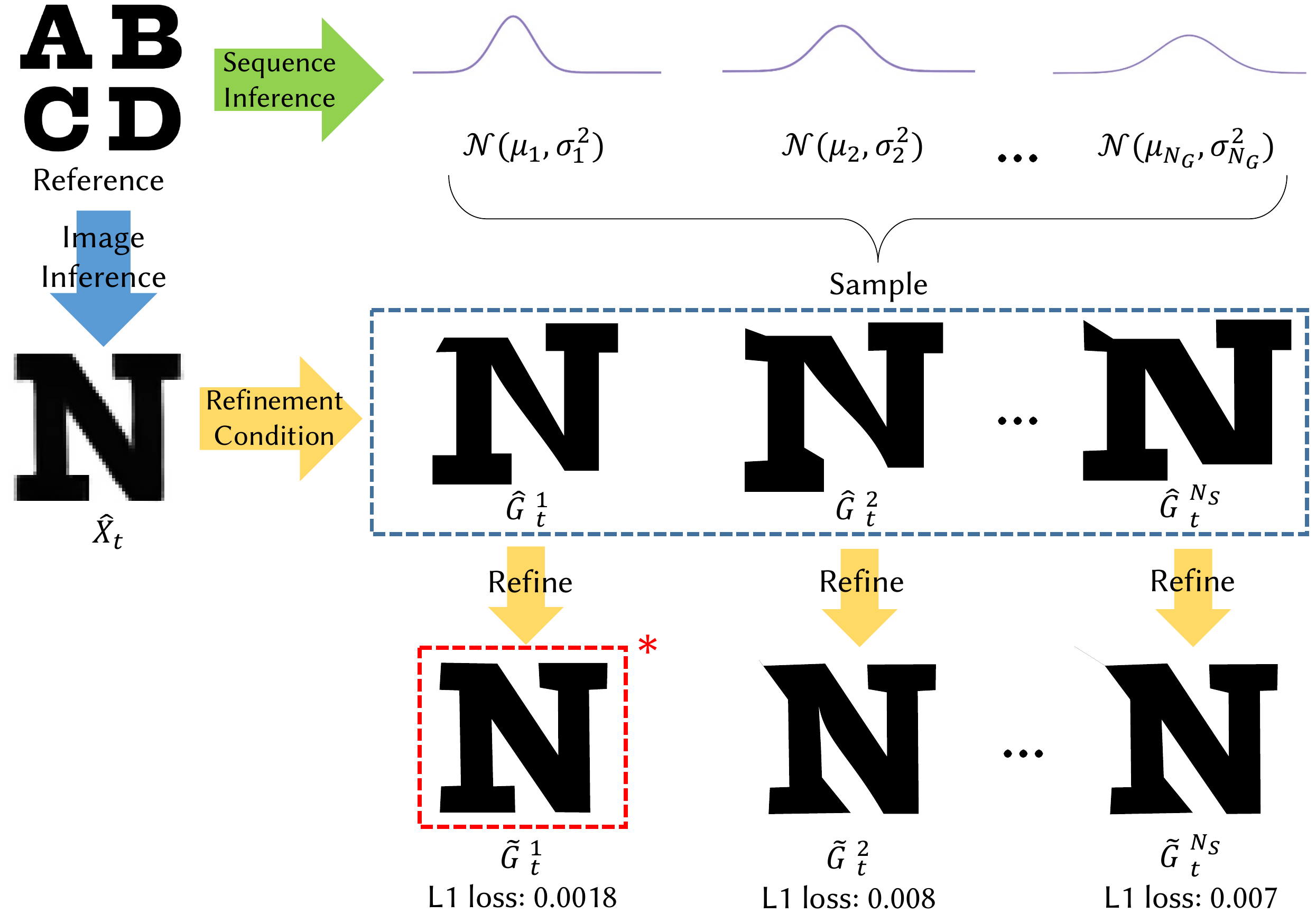}
  \caption{The pipeline of our further refinement operation in the inference stage. We sample a number of vector glyphs from the predicted distributions, which are then further refined according to the generated images.}
  \label{fig:FurtherRefinement}
\end{figure}

Fig.~\ref{fig:FurtherRefinement} demonstrates how we perform further refinement on the generated glyphs in the inference stage.
First, we sample a number of candidate glyphs from the output mixture distributions:
\begin{equation}
\hat{G}_{t}^{1},\hat{G}_{t}^{2},...,\hat{G}_{t}^{N_{s}} \sim GMM(\{(\lambda_{k},\mu_{k}, \sigma_{k}) | 1\leq k \leq N_{G}\}).
\end{equation}
The reason of implementing multiple sampling is that the most probable candidate sampled based on the greedy strategy (choosing the distribution with the maximum probability at each time step) is not necessarily the best one.
This phenomenon happens in NLP tasks, where beam search has been proposed to handle the problem.
However, the searching space in our task is infinite unlike the discrete spaces in NLP tasks, and thus beam search is unsuited for this task.
Therefore we sample a fixed number of plausible candidates and then find the best one from them.
As described in Section 3.1, each glyph $\hat{G}_{t}^{\theta}$ ($1\leq \theta \leq N_{s}$) is composed of command types $\hat{z}_{t}^{\theta}$ and command coordinates $\hat{p}_{t}^{\theta}$:
\begin{equation}
\hat{G}_{t}^{\theta} = (\hat{z}_{t}^{\theta},\hat{p}_{t}^{\theta}).
\end{equation}
We merge the starting point and ending point of each path to make the path closed, since they are both sampled from Gaussian Distributions and typically not identical.
Then, we utilize the differentiable rasterizer proposed by ~\cite{li2020differentiable}, denoted as $F_{ras}^{m}$, to further refine our generated vector glyphs.
Specifically, we fix the command types $\hat{z}_{t}^{\theta}$ and fine-tune the coordinates values $\hat{p}_{t}^{\theta}$ to minimize the L1 distance between the rasterization result and the synthesized glyph image $\hat{X}_{t}$:
\begin{equation}
 \tilde{p}_{t}^{\theta} = \mathop{\arg\min}_{\hat{p}_{t}^{\theta}} ||F_{ras}^{m}(\hat{z}_{t}^{\theta}, \hat{p}_{t}^{\theta}) - \hat{X}_{t}||_{1},
\end{equation}
where the optimization is performed by the gradient descent (GD) algorithm.
Afterwards, we obtain the refined vector glyphs which are denoted as:
\begin{equation}
  \tilde{G}_{t}^{\theta} = (\hat{z}_{t}^{\theta},\tilde{p}^{\theta}_{t}).
\end{equation}
We select the best result from all refined glyphs as the final output, whose rendered image has the minimum L1 distance with the corresponding synthesized glyph image:
\begin{equation}
  \theta^{*} = \mathop{\arg\min}_{\theta} ||F_{ras}^{m}(\tilde{G}_{t}^{\theta}) - \hat{X}_{t}||_{1}.
\end{equation}
As we can see from Fig.~\ref{fig:FurtherRefinement}, the refinement performance is significantly affected by the synthesized initial glyphs. If the number and types of drawing commands of these initial glyphs are similar to the ground truth, most probably visually-pleasing vector glyphs will be obtained after our refinement procedure. 
This refinement process might cause nonadjacent segments to have intersections which bring small artifacts on the output glyphs. Therefore, after implementing refinement, we also detect these unexpected intersections and delete the parts within them. In this manner, most of those artifacts can be automatically removed.
\section{Experiments}
\subsection{Implementing Details and Dataset}
The number of reference glyphs $N_{r}$ is set to 4 and we also conduct experiments to test the performance of our model with different values of $N_{r}$.
The number of potential Gaussian Distributions $N_{G}$ is set to 50.
The number of sampled glyphs $N_{s}$ in the inference stage is set to 20.
We utilize Adam~\cite{kingma2015adam} as the optimizer of 
our model.
The image resolution is set to $64\times64$ in the training stage.
To obtain better refinement results in the inference phase, we employ a U-Net~\cite{ronneberger2015u} based image super-resolution model to increase the resolution of initial synthesis images from $64\times64$ to $256\times256$.
\par
For fair comparison, we employ the SVG-Fonts Dataset introduced by SVG-VAE~\cite{lopes2019learned} in our experiments.
A subset of SVG-Fonts is used (around 8K fonts for training and 1.5K fonts for testing) since the size of the original dataset is too huge (around 14M fonts).
Although the amount of training fonts we utilize is much smaller than SVG-VAE, our model still achieves significantly better performance than SVG-VAE.
\subsection{Performance of Our Neural Differentiable Rasterizer}
\begin{figure}[t!]
  \centering
  \includegraphics[width=\columnwidth]{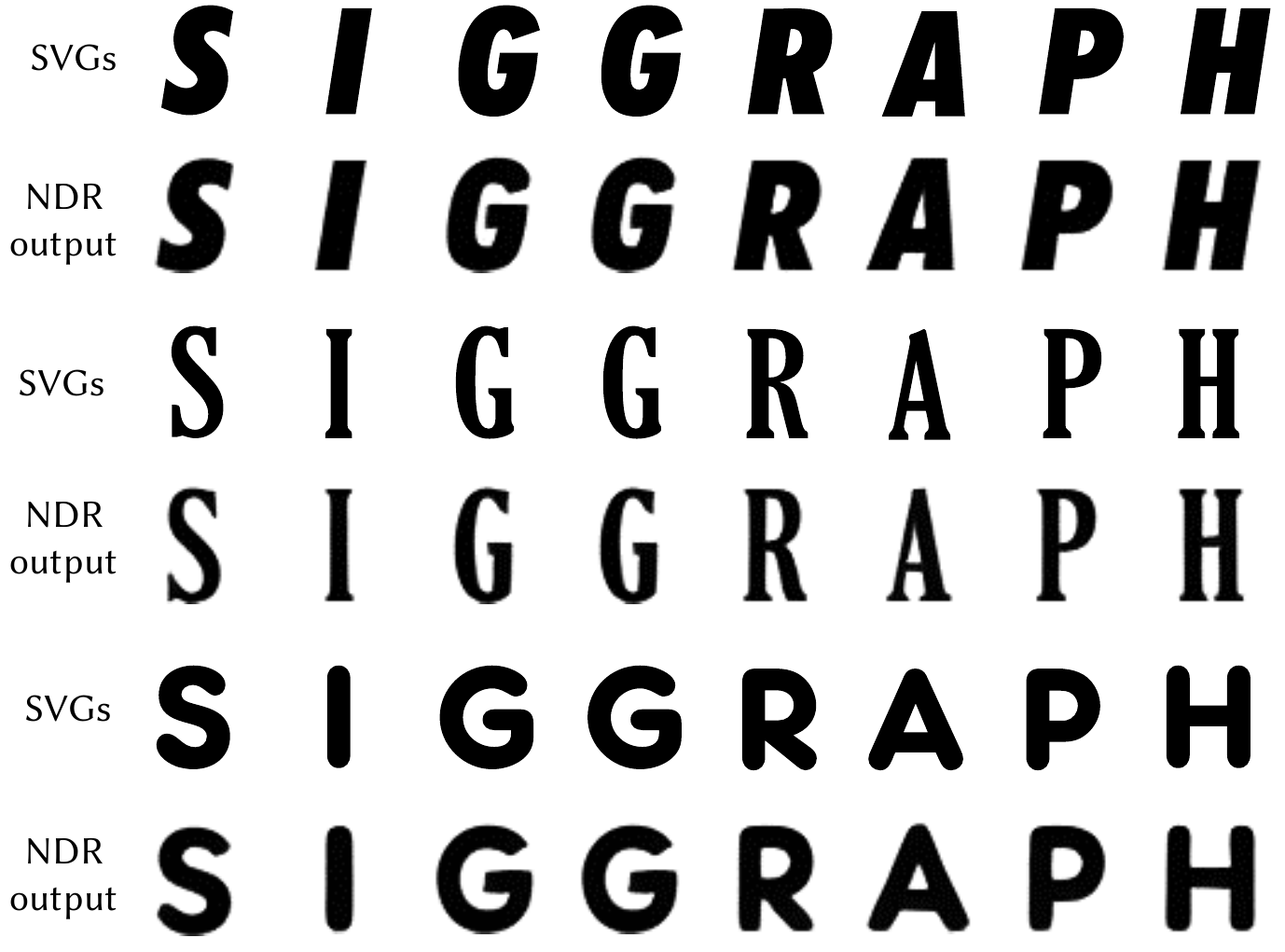}
  \caption{The performance of our pretrained Neural Differetiable Rasterizer (NDR). These input SVGs are from the testing dataset.}
  \label{fig:NeuralRasterizerRes}
\end{figure}
\begin{figure}[t!]
  \centering
  \includegraphics[width=\columnwidth]{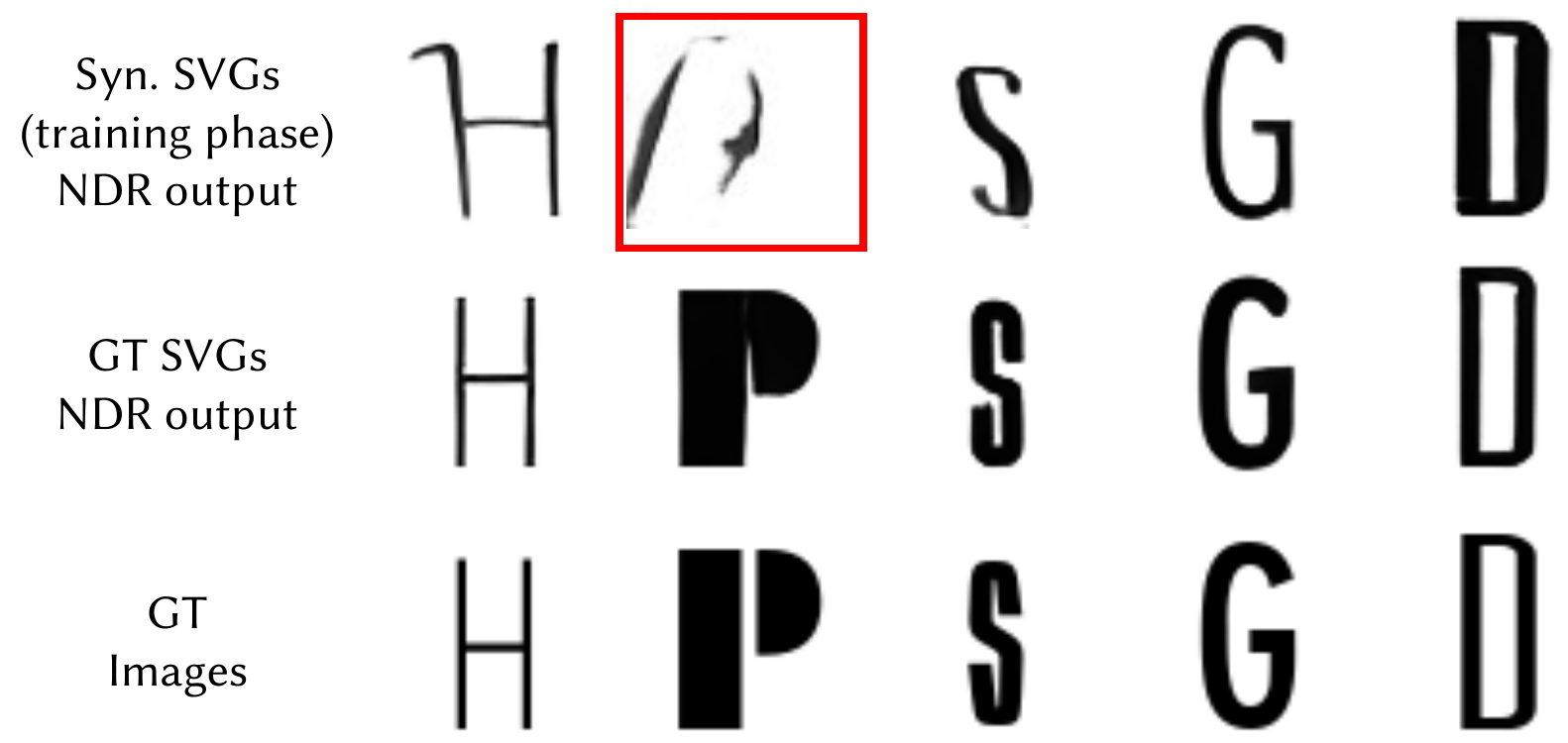}
  \caption{What NDR outputs when we feed it with $\hat{G}_{t}$ (synthesized drawing commands during training). The red rectangle marks out an example of NDR outputs when it receives an invalid sequence. ``Syn.'' denotes synthesized and ``GT'' denotes ground-truth.}
  \label{fig:NeuralRasterizerResTrain}
\end{figure}
After being trained with a large amount of data, our Neural Differentiable Rasterizer (NDR) manages to learn the mapping function from a given SVG file's drawing commands to its corresponding raster image.
As shown in Fig.~\ref{fig:NeuralRasterizerRes}, our trained NDR is able to convert the input SVGs into raster images that only contain negligible differences in local details compared to the ground truth.
In Fig.~\ref{fig:NeuralRasterizerResTrain}, we also show what NDR outputs when it is fed with randomly sampled drawing commands $\hat{G}_{t}$ during training.
The red rectangle demonstrates the NDR output when it receives an invalid sequence, which lacks an ending command.
This situation cannot be properly handled by the differentiable rasterizer proposed by~\cite{li2020differentiable} because it needs closed outlines.
We then plug our NDR into the main model for training, and we also conduct ablation studies to verify its effectiveness.


\subsection{Parameter Studies}
Here, we conduct experiments to investigate how the number of glyphs fed into the style encoder (i.e., $N_{r}$) affects our model's performance.
The results are shown in Table~\ref{tab:ParameterNr}, from which we can observe that generally larger $N_{r}$ results in better performance.
However, larger values of $N_{r}$ not only increase the computational cost of our model, but also take more times for designers to manually create more input vector glyphs.
When $N_{r} > 4$, the performance improvement brought by the increment of $N_{r}$ becomes marginal.
To achieve a good balance between the model size and performance, we choose $N_{r}$ = 4 as the default setting of our model unless otherwise specified.
\begin{table}
	\centering
	\caption{How the setting of $N_{r}$ affects our model's performance. The loss values are calculated on the validation dataset after our model finishes training.}
	\begin{tabular}{lccc}
		\toprule
		$N_{r}$      & ImgRec Loss $\downarrow$    & MDN Loss $\downarrow$  & CE loss $\downarrow$ \\
		\midrule
		1     &          0.2145  & -1.0895   & 0.1056 \\
		2    &          0.1536 & -1.1948   & 0.0987 \\
		4      &         0.1478  & -1.3071  & 0.0912  \\
		8   & \textbf{0.1389} & \textbf{-1.3749} & \textbf{0.0879} \\
		\bottomrule
	\end{tabular}
    \label{tab:ParameterNr}
\end{table}
\subsection{Ablation Study}
For the purpose of analyzing the impacts of different
modules, we conduct a series of ablation studies by removing or changing the proposed modules in our model.
\subsubsection{Quantitative Experiments}
\begin{figure}[t!]
  \centering
  \includegraphics[width=\columnwidth]{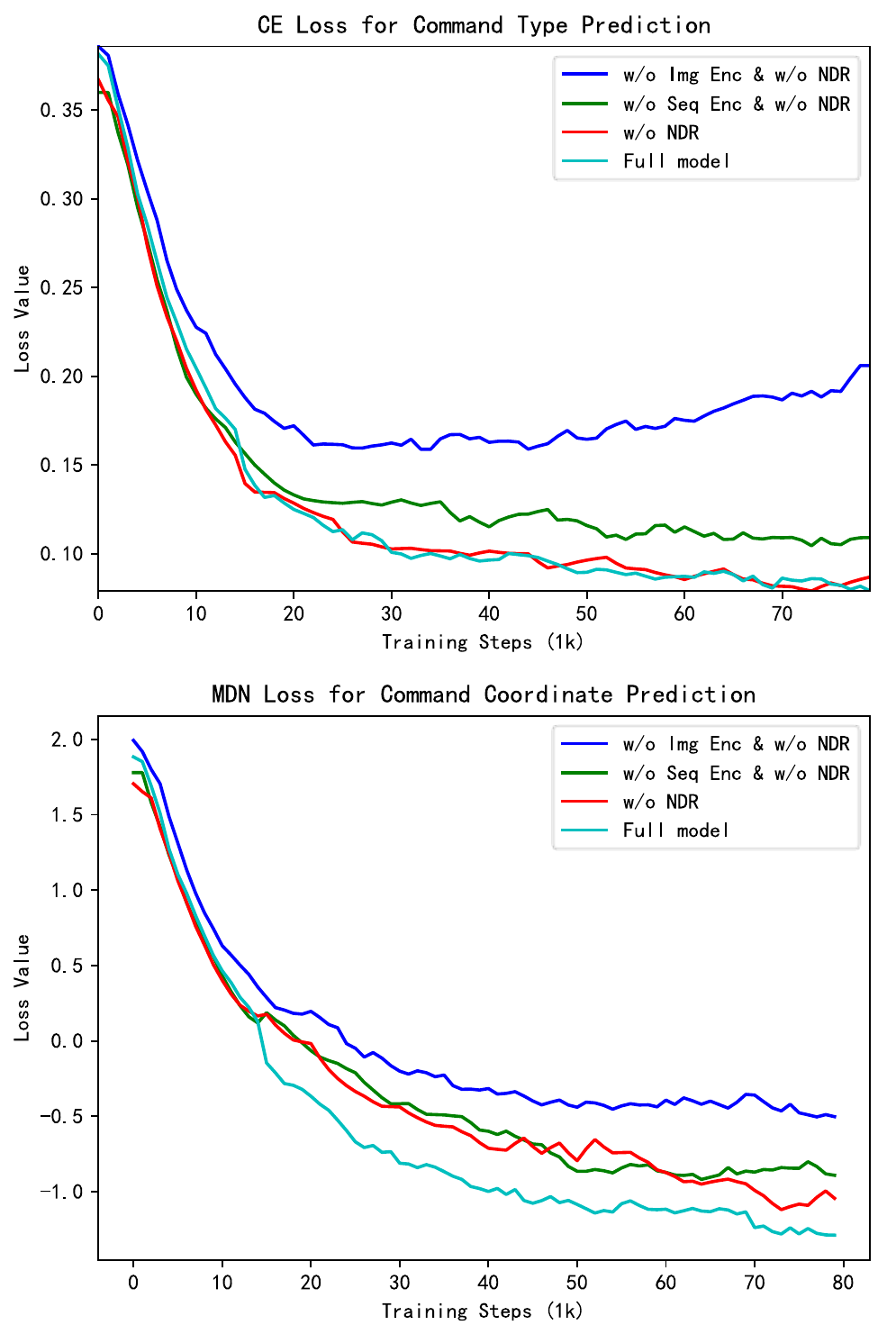}
  \caption{The loss curves of our model evaluated on the validation dataset when trained with or without the CNN Encoder, the LSTM Encoder and NDR. ``Img Enc'' and ``Seq Enc'' denote Image Encoder and Sequence Encoder, respectively. ``NDR'' denotes Neural Differentiable Rasterizer.}
  \label{fig:AblationStudyQuan}
\end{figure}
The loss curves of $L_{CE}$ and $L_{MDN}$, which can be found in Fig.\ref{fig:AblationStudyQuan}, demonstrate how the dual-modality fusion strategy promotes the model's performance of reconstructing the target vector glyphs.
We witness a notable improvement brought by the proposed dual-modality learning strategy and NDR module.
More specifically, encoding only the sequence modality results in the worst performance (blue curves).
Encoding only the image modality results in better performance (green curves), but not so good as encoding both modalities (red curves), especially in the command type prediction.
This is because the image modality cannot precisely represent local detailed structures, which makes the model struggle to discriminate between lines and curves.
Fusing the features of both modalities helps to solve this issue (examples can be found in Fig.~\ref{fig:AblationStudyQual}).
Since NDR is deployed to further align the generated vector glyphs with the target glyph images, it mainly contributes to the performance of coordinate prediction.
\subsubsection{Qualitative Experiments}
\begin{figure}[t!]
  \centering
  \includegraphics[width=\columnwidth]{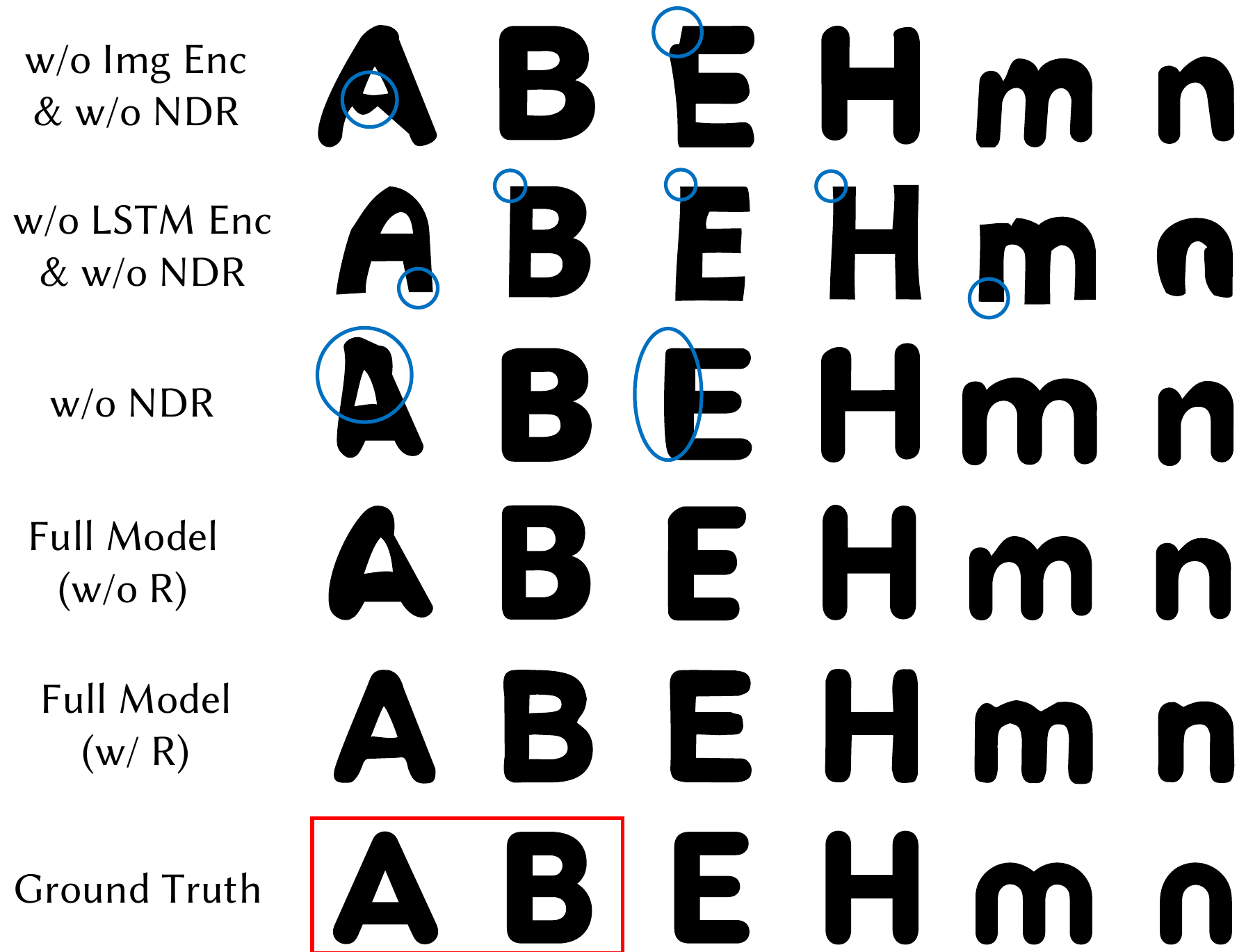}
  \caption{The performance of our model under different configurations. The input reference glyphs are marked by red rectangles. "w/o R" denotes the method that synthesizes raw vector glyphs without further refinement. The blue circles highlight the shortcomings of each incomplete configuration of our model.}
  \label{fig:AblationStudyQual}
\end{figure}
We also provide qualitative experimental results to verify the effectiveness of each proposed module, which are shown in Fig.~\ref{fig:AblationStudyQual}.
Without the Image Encoder, our model tends to generate some redundant lines and curves with severe distortions.
Without the Sequence Encoder (i.e., LSTM Encoder), our model fails to reconstruct the local characteristics of fonts (e.g., round corners) in the reference inputs.
The dual-modality fusion strategy successfully helps our model sufficiently learn the font style features.
Also, our proposed NDR alleviates the location shift problem and helps to construct vector glyphs with higher quality.
\subsection{Few-shot Vector Font Generation}
\begin{figure*}[t!]
  \centering
  \includegraphics[width=16cm]{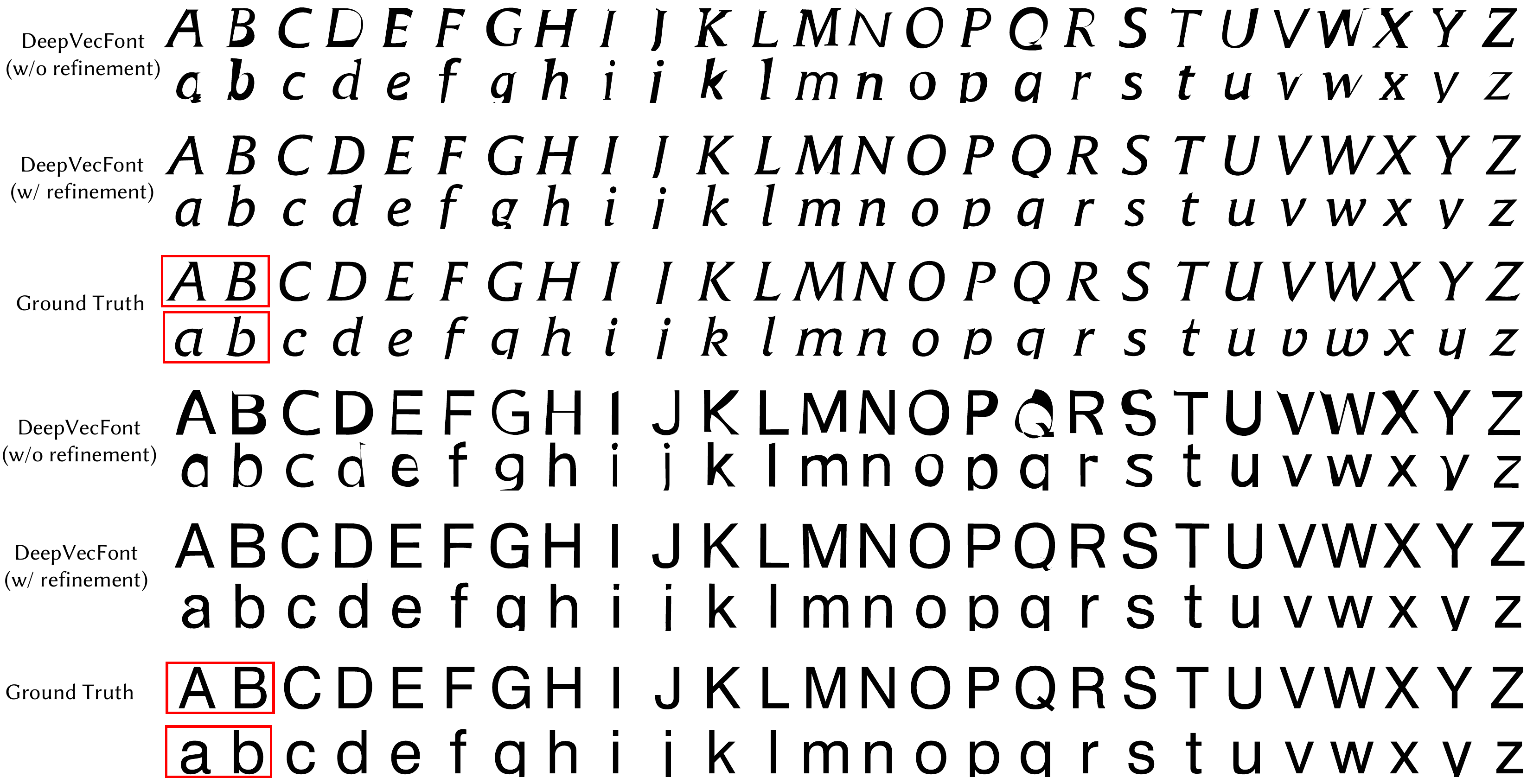}
  \caption{Generating the whole vector font given a few reference glyphs in the testing dataset. The input reference glyphs are marked by red rectangles. More results are shown in the supplemental material.}
  \label{fig:FewShotFontGeneration1}
\end{figure*}
\begin{figure}[t!]
  \centering
  \includegraphics[width=\columnwidth]{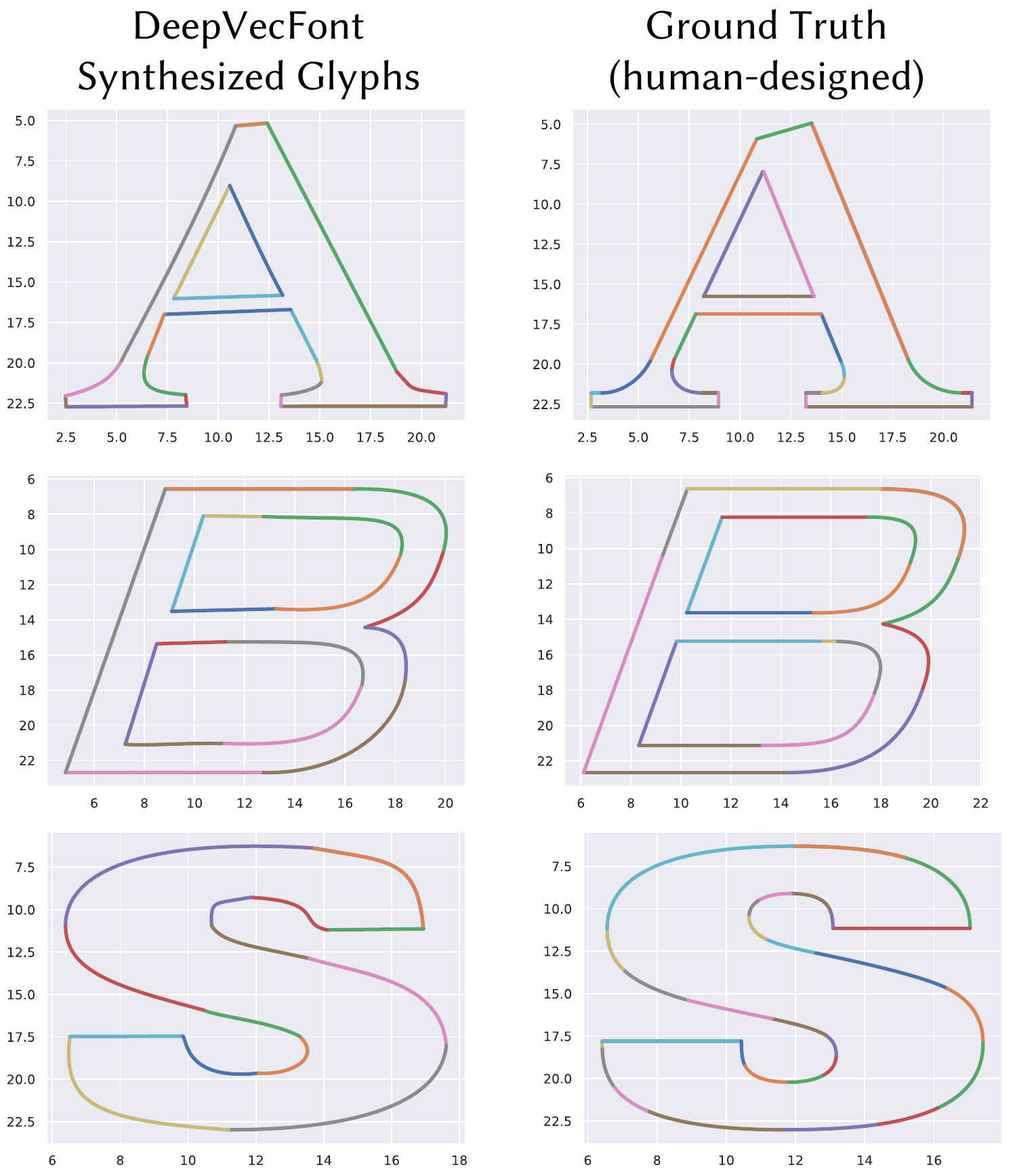}
  \caption{A detailed illustration of our synthesized vector glyphs. Please zoom in for better inspection. We use different colors (a periodic color sequence i.e., orange, green, red, ..., blue) to distinguish between different draw-commands.}
  \label{fig:ShowGraphics}
\end{figure}
As shown in Fig.~\ref{fig:FewShotFontGeneration1}, we demonstrate the effectiveness of our method in the task of generating all other glyphs by giving a few ($N_{r}$) reference glyphs.
The input reference glyphs are marked by red rectangles in Fig.~\ref{fig:FewShotFontGeneration1}, from which we can see that our model is capable of synthesizing the whole vector font given just a small number of reference glyphs. Especially for the second font shown in Fig.~\ref{fig:FewShotFontGeneration1}, almost all glyphs in the synthesized vector font are approximately identical to the human-designed glyphs. 
Yet, we also observe that, for the first font shown in Fig.~\ref{fig:FewShotFontGeneration1}, many local details of our synthesis results are slightly different against the ground-truth glyphs (especially in the upper cases).
Considering that our model only receives the glyphs of `A', `B', `a' and `b' as reference inputs, our synthesized glyphs of other characters have already sufficiently embodied the style feature of these input samples. More importantly, the style of glyphs in the same synthesized vector font is consistent, and they all look visually pleasing.\par

In Fig.~\ref{fig:ShowGraphics}, we demonstrate the detailed graphics of DeepVecFont's synthesized glyphs and ground-truth glyphs.
A periodic color sequence (from orange to blue) is used to mark different draw-commands.
We can see that the ground-truth (human-designed) fonts tend to use some unnecessary segments and the starting point of a path cannot be precisely predicted, which makes the task very challenging.
On the contrary, DeepVecFont is capable of reconstructing outlines that are as compact as possible.\par

\begin{figure}[t!]
  \centering
  \includegraphics[width=\columnwidth]{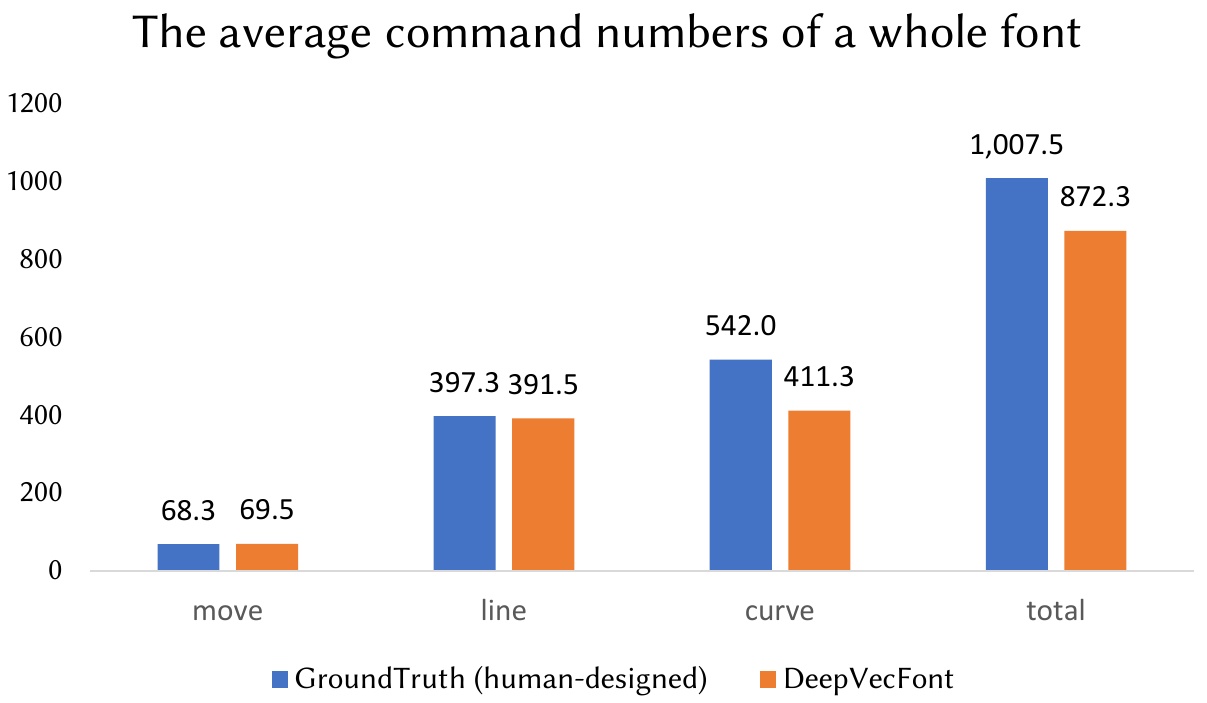}
  \caption{The average command numbers of our DeepVecFont's synthesized fonts and the corresponding human-designed fonts selected from the testing dataset. The drawing commands of vector fonts synthesized by our method are slightly less than the human-designed fonts.}
  \label{fig:AvgCommandNumbers}
\end{figure}

In Fig.~\ref{fig:AvgCommandNumbers}, we show the average command numbers of a whole font synthesized by our method and compare them with the ground truth (i.e., human-designed vector fonts). As we can see, DeepVecFont has no difficulty in predicting the ``move'' commands, which determine the number of paths in a vector glyph.
Furthermore, DeepVecFont prefers to use less cubic Bézier curves than human designers, which has two possible reasons:
(1) Human designers occasionally use unnecessary cubic curves when designing vector glyphs, such as the cases in Fig.~\ref{fig:WhyMDNNeeded} and Fig.~\ref{fig:ShowGraphics}.
(2) Many local stylistic details are constructed by curves. Given just a small number of reference glyphs, DeepVectFont cannot always perform perfect reconstructions.
\subsection{Vector Font Interpolation}
Benefiting from the normalization training of latent spaces, our model is able to perform vector font interpolation between different fonts by manipulating their latent codes.
Assume that we have two fonts denoted as the font $a$ and the font $b$, the interpolated feature of them can be computed by:
\begin{equation}
f_{ip}  = (1 - \lambda)\cdot f(a) + \lambda \cdot f(b).
\end{equation}

\begin{figure}[t!]
  \centering
  \includegraphics[width=\columnwidth]{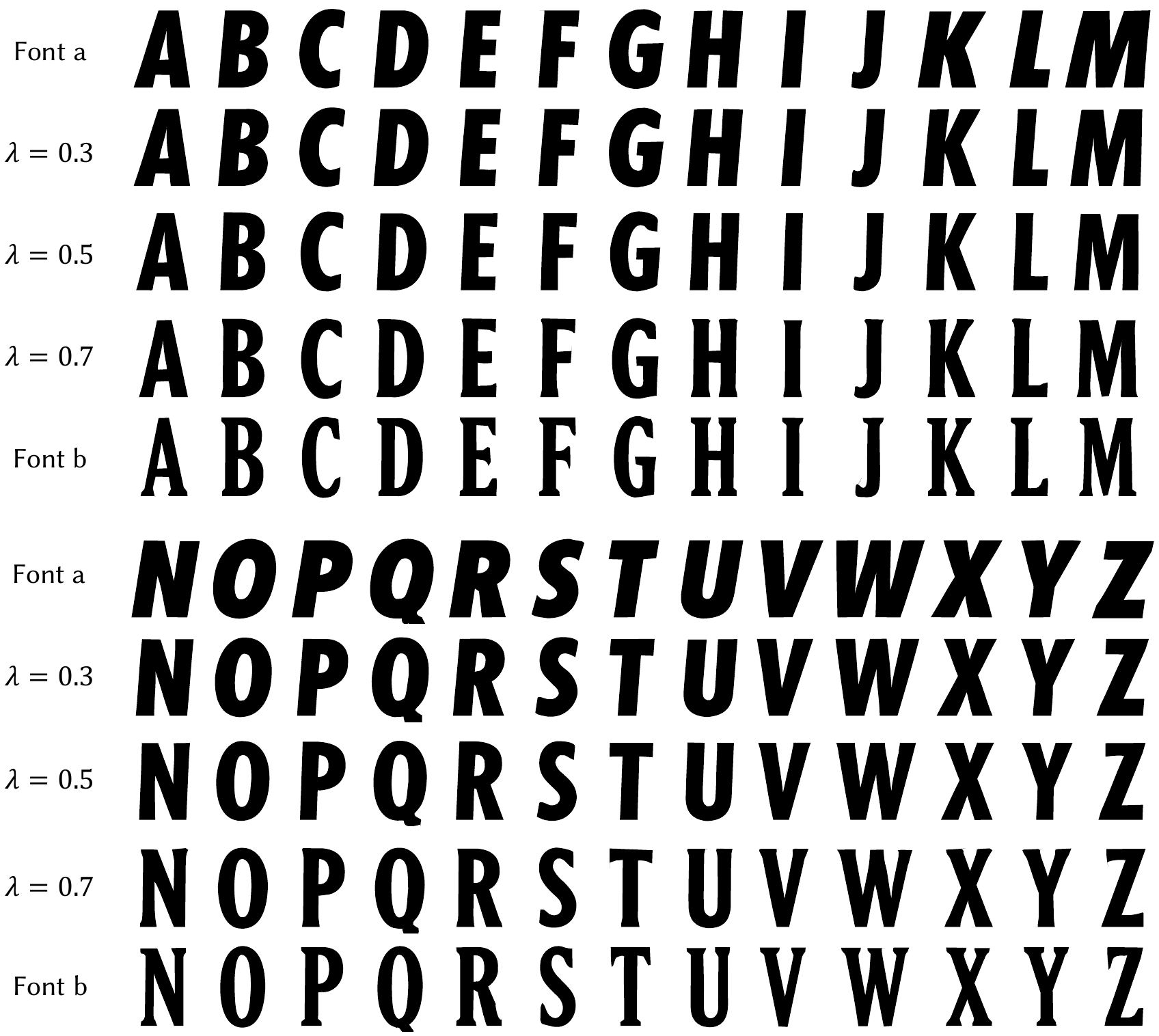}
  \caption{Interpolations between two different vector fonts, where the weight, italic and serif change smoothly.}
  \label{fig:InterpPart1}
\end{figure}

We then send $f_{ip}$ into the decoders of our model and obtain the interpolated glyph images and vector glyphs.
Afterwards, the refinement operation is conducted on the interpolated vector glyphs.
Fig.~\ref{fig:InterpPart1} shows that our model achieves smooth interpolation between different fonts and is capable of automatically generating visually-pleasing vector fonts with brand-new styles.
As we can see from Fig.~\ref{fig:InterpPart1}, the italic and weight decline while the serif appears gradually during the interpolation process. Despite the obvious difference between every pair of font styles, all glyphs transform smoothly and naturally from the source font style to the target. 
\subsection{The Further Refinement Process}
\begin{figure}[t!]
  \centering
  \includegraphics[width=\columnwidth]{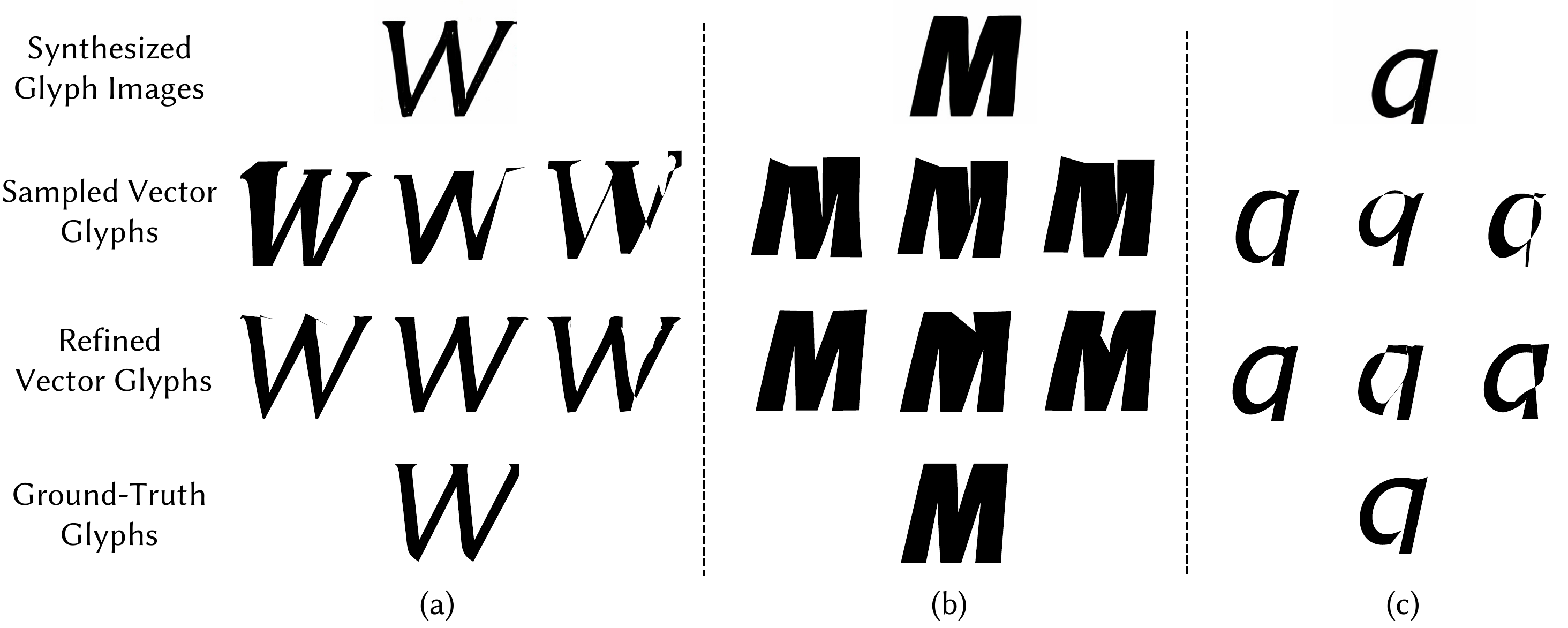}
  \caption{More examples of the refinement process in the reference stage. In each case, we demonstrate three candidate vector glyphs whose probabilities of being sampled decrease from left to right.}
  \label{fig:RefineSeachingExamples}
\end{figure}

We show more examples of the refinement process in Fig.~\ref{fig:RefineSeachingExamples}.
Three important conclusions can be drawn through our experiments:
(a) Glyphs with the highest probabilities of being sampled are not necessarily the best candidates, which verifies the necessity of our multiple sampling strategy. (b) The refinement process is stochastic, to a certain degree, on account of the gradient descent algorithm. Three similar initial samples have different refined results. (c) The crossing of outline segments often leads to severe distortions on the refinement results.
\begin{figure*}[t!]
  \centering
  \includegraphics[width=\textwidth]{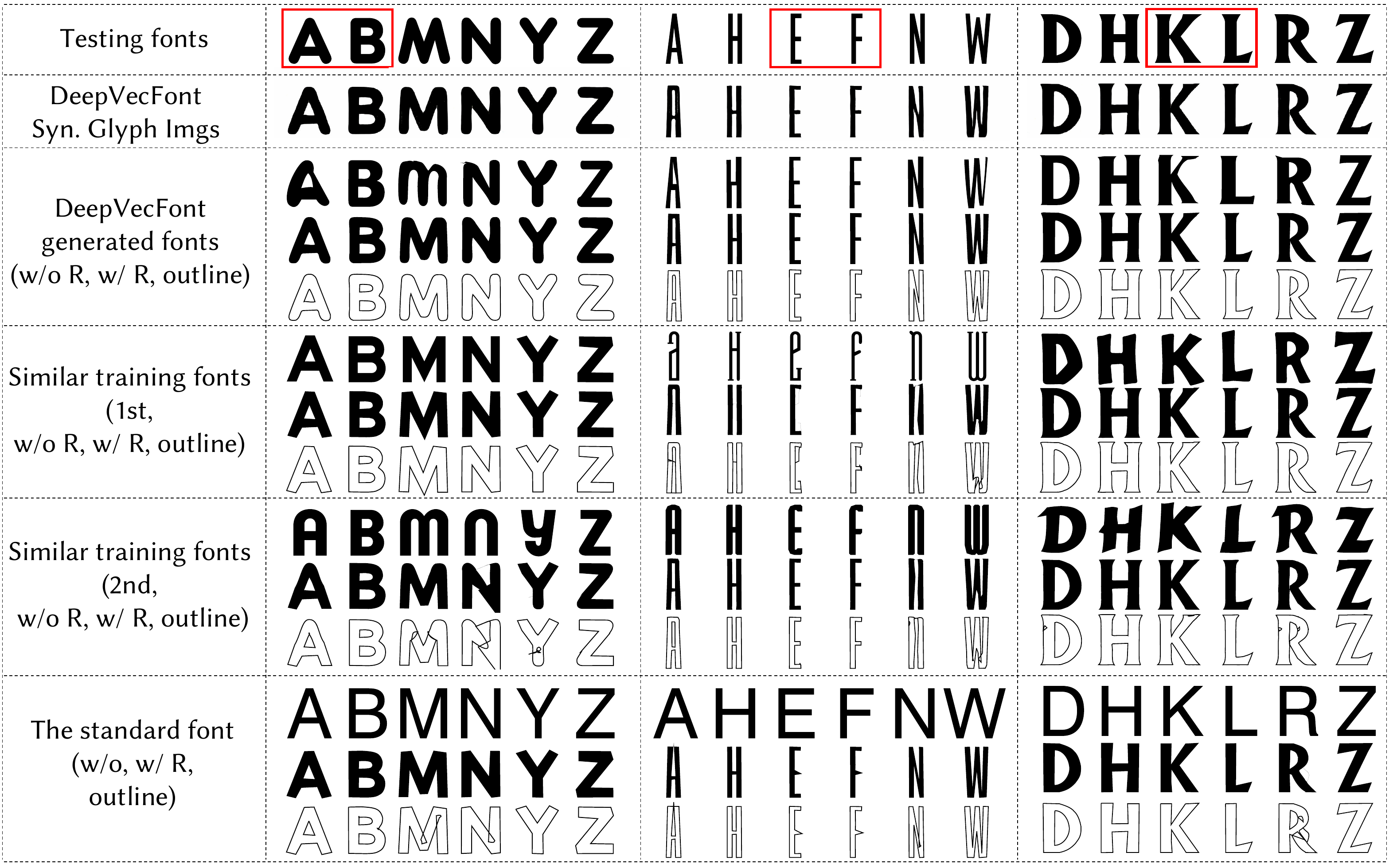}
  \caption{Comparison of synthesized vector fonts obtained by our method and the approach that directly conducts refinements on the most similar training fonts and the standard font. The input reference glyphs are marked by red rectangles ($N_{r} = 2$ for each case). "Syn. Glyph Imgs" denote synthesized glyph images. "w/o R" and "w/ R" denote without and with refinement, respectively. In this figure, we show the refinement results from the top-2 most similar fonts for each case.}
  \label{fig:RefiningFromSimilarFonts}
\end{figure*}

\subsection{Refining the Most Similar Fonts}
To obtain the final vector fonts, we further refine the initially generated vector fonts according to their corresponding generated images.
Intuitively, there exists another approach that is to directly refine the glyphs with the most similar style as the input samples or a chosen standard font style in the training dataset.
We also compare our method with this approach and the results are shown in Fig.~\ref{fig:RefiningFromSimilarFonts}.
The training font $p^{*}$ whose style is most similar to a given testing font $q$ is retrieved by the following equation:
\begin{equation}
  p^{*} =  \mathop{\arg\min}_{p}|| f(p) - f(q)||_{2},
\end{equation}
where $f(p)$ and $f(q)$ are the latent codes of the font $p$ and $q$, respectively.
We select Arial as the standard font in this experiment.
From Fig.~\ref{fig:RefiningFromSimilarFonts}, we can see that results refined from DeepVecFont's synthesized fonts significantly outperform those refined from the most similar training fonts and Arial.
For example,
the first font shares similar smooth corners with the second most similar training font but their typologies are quite different.
It is difficult to refine fonts with very different typologies into our desired fonts.
The second font has an identical topology with Arial but their aspect ratios are quite different, which makes the GD algorithm stuck in the local optima.
As a result, the refined font from Arial occasionally has artifacts in glyphs such as `A', `E' and `F'.
For the third font, its similar training fonts have too many redundant commands, which becomes wobbles and dents in the glyphs of the refined fonts.  

\subsection{Random Sampling From Latent Space}

\begin{figure}[t!]
  \centering
  \includegraphics[width=\columnwidth]{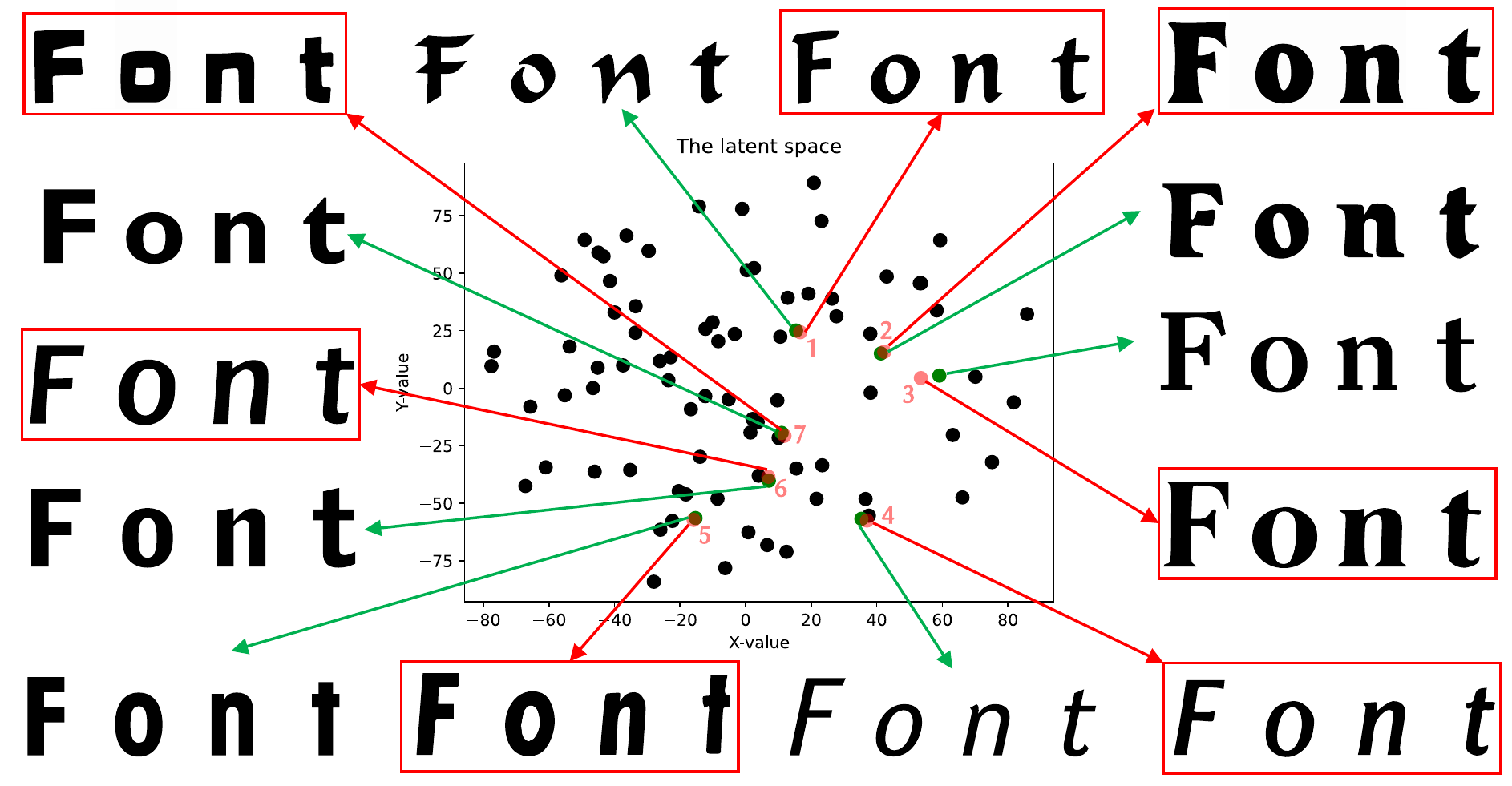}
  \caption{Random generation from the latent space. The latent space is visualized by T-SNE. The pink (red with 50\% opacity) points denote the randomly generated fonts (marked by red rectangles). The green points denote the training fonts that are the most similar to these generated fonts. The black points denote other training fonts. Please zoom in for better inspection.}
  \label{fig:RandomGeneration}
\end{figure}

Benefiting from the united and normalized latent space, our model can easily synthesize fonts in new styles. Specifically, we randomly sample 128-dimensional vectors from the Gaussian Distribution $\mathcal{N}(0,I)$ as the latent codes for our model to generate glyph images and initial vector glyphs.
Following the procedure mentioned above, the vector glyphs are then further refined according to the synthesized glyph images.
In Fig.~\ref{fig:RandomGeneration}, we demonstrate some randomly generated fonts (refined) and mark them in the latent space visualized by T-SNE~\cite{van2008visualizing}.
It can be observed that our generated fonts are visually pleasing and different from the most similar training fonts.

\subsection{Comparison with Other Methods}
\begin{figure*}[t!]
  \centering
  \includegraphics[width=\textwidth]{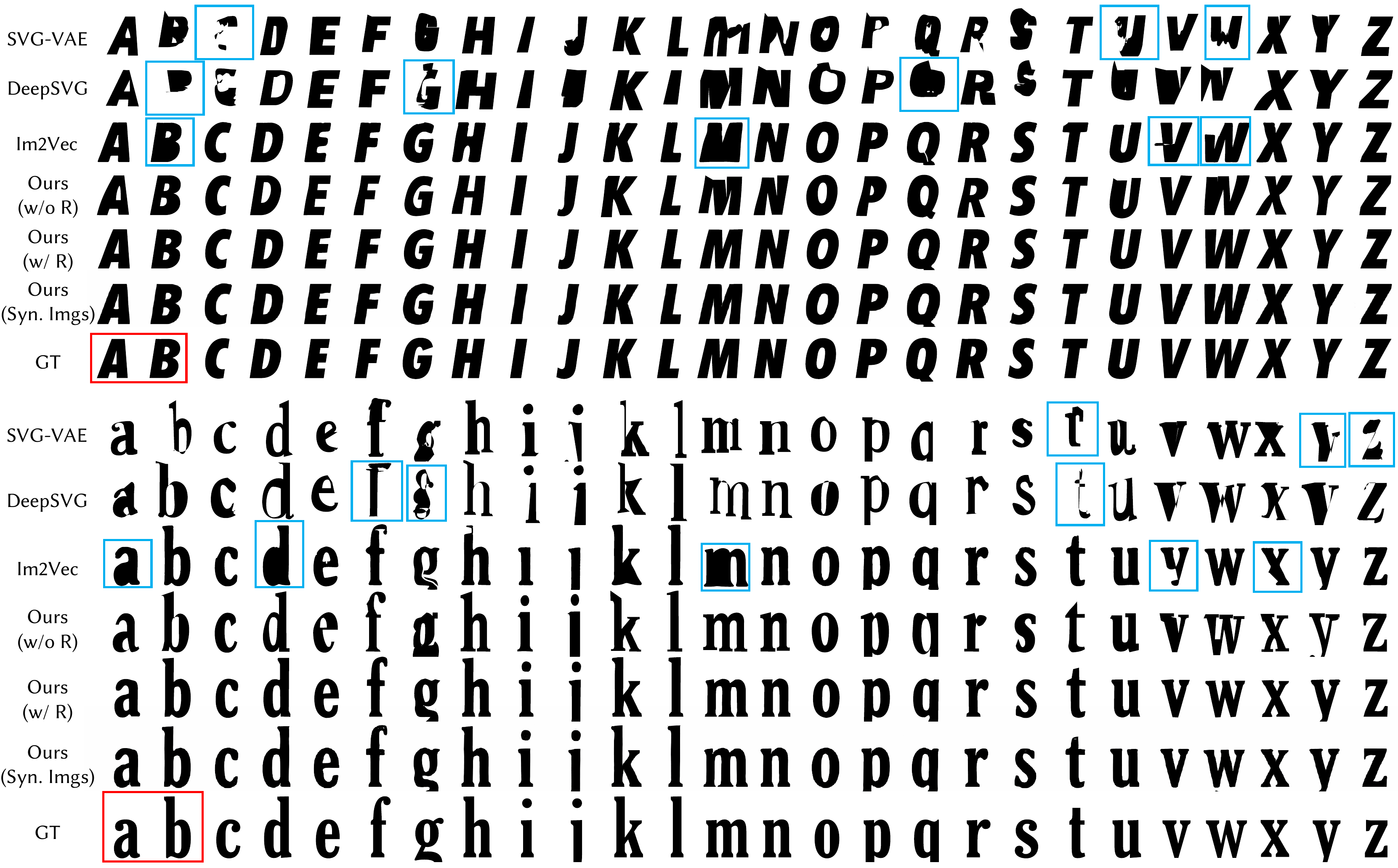}
  \caption{Few-shot vector font generation results of SVG-VAE, DeepSVG, Im2Vec and our method. "w/o R" and "w/ R" denote vector glyphs "without refinement" and "with refinement", respectively. "Syn. Imgs" denotes the synthesized images of our method, which are the input images of Im2Vec. The input reference glyphs are marked by red rectangles (other two reference glyphs are not shown in this figure, which are `a', `b' and `A', `B' for each case, respectively). The blue rectangles highlight the representative failure cases generated by existing models.}
  \label{fig:FewShotFontGenerationComparison}
\end{figure*}
We compare our method with three recently-proposed approaches: SVG-VAE~\cite{lopes2019learned}, DeepSVG~\cite{carlier2020deepsvg} and Im2Vec~\cite{reddy2021im2vec}.
It should be pointed out that the Im2Vec approach compared here takes the synthesized glyph images of our method as input to obtain the vector glyphs.
Since SVG-VAE and DeepSVG learn the font style feature from a single glyph, we send our $N_{r}$ reference glyphs into their models separately and then compute the mean value of the extracted style features.
Since the synthesis results obtained by DeepSVG lack the necessary information for contour filling, we first manually process them and then render their normal raster glyph images for clear comparison in our experiments.
From Fig.~\ref{fig:FewShotFontGenerationComparison} we can see that DeepSVG and SVG-VAE often tend to generate vector glyphs with severe distortions and large amounts of artifacts, which consist of numerous curves.
For Im2Vec, there exist three main drawbacks:
(1) it prefers to first fit the big outlines but sometimes ignores the small ones in glyphs (e.g., `B', `a' and `d');
(2) it tends to be stuck in local optimums for glyphs with multiple concave regions (e.g, `m', `M' and `k');
(3) the lines of glyphs are not well fitted and the curves are not as smooth as ours.
Thereby, a conclusion can be made that the details of vector fonts cannot be reconstructed well with only image supervision.
The accurate prediction of drawing commands for curves is very difficult because there exist six arguments to be regressed. Thanks to the duel-modality learning strategy adopted in our method, the quality of our synthesized vector fonts is already significantly better than those obtained by those existing approaches even without refinement, and is comparable to the human-designed ground truth after implementing our refinement procedure.\par
\begin{figure}[t!]
  \centering
  \includegraphics[width=\columnwidth]{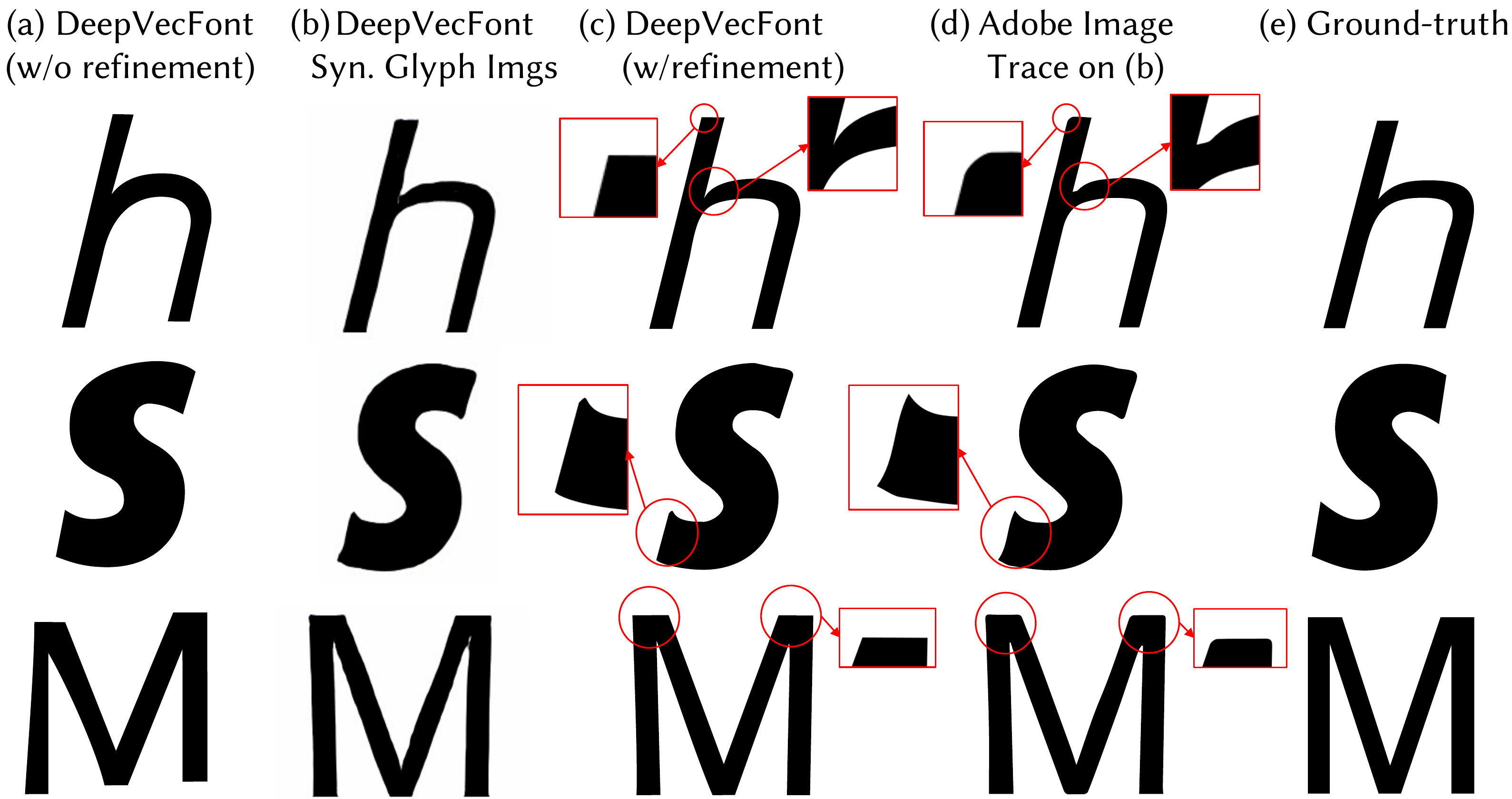}
  \caption{Comparisons between our method and Adobe Image Trace. Please zoom in for better inspection.}
  \label{fig:ComparisonWithAIT}
\end{figure}

%
We also compare our proposed DeepVecFont with Adobe Image Trace (AIT) by sending our synthesized high-resolution (i.e., $256 \times 256$) images into AIT.
As shown in Fig.~\ref{fig:ComparisonWithAIT}, there exist two problems in the AIT's rasterization of our generated images:
(1) AIT fits each contour of images by using high-order curves, even for lines;
(2) Many flaws on our generated images are preserved due to the requirement of precise curve fitting.
On the contrary, DeepVectFont can synthesize glyphs smoothly mixed with curves and lines since it outputs raw (without refinement) vector glyphs with accurate command types.
Therefore, as shown in Fig.~\ref{fig:ComparisonWithAIT}, most flaws from synthesized images can be avoided by our DeepVecFont but are incorrectly preserved in the results obtained by AIT.
This is mainly because the length and types of correctly-predicted drawing commands in each vector glyph are fixed in our refinement procedure.
In Table~\ref{tab:comparison-para-prefer},
We also conduct quantitative experiments to compare the performance of our DeepVecFont with other existing methods or commercial software: AIT, SVG-VAE and DeepSVG, which further demonstrates the superiority of our DeepVecFont compared to the state of the art.
The reconstruction errors are measured by computing the average L1 distances between the rasterized synthesis results of different methods and the corresponding ground-truth glyph images (at the resolution of $256 \times 256$) in the testing dataset.
\begin{table}
	\centering
	\caption{Comparison of reconstruction errors (L1 losses) for different vector font generation methods. ``DVF'' denotes DeepVecFont. ``DVF (RFS)'' denotes DVF's results by refining from the most similar fonts.}
	\begin{tabular}{lc}
		\toprule
		Model      &  Reconstruction Error $\downarrow$    \\
		\midrule
		AIT     &          0.031     \\
		SVG-VAE    &  0.090    \\
		DeepSVG     &         0.120     \\
		DVF (RFS)    & 0.058  \\
		\textbf{DVF}    & \textbf{0.021}  \\
		\bottomrule
	\end{tabular}
    \label{tab:comparison-para-prefer}
\end{table}
\subsection{User Study}
We invite people from different occupations and ages to evaluate our synthesized fonts. The participates consist of 11 professional font designers and 20 non-designers (including students, teachers, company employees, etc). First, we conduct a Turing test where the participates are asked to discriminate between our synthesized fonts and human-designed fonts. For each font, we provide a list of 52 vector glyphs (`A'-`Z' and `a'-`z') where half of them are our synthesized results and the others are human-designed glyphs. The participates are asked to select the glyphs which they think are generated by machines (AI). They are allowed to zoom in to inspect the details of each glyph. There are 6 fonts with different styles in this test. After finishing the Turing test, all participates are asked to give ratings of our synthesized fonts in terms of quality and style consistency, respectively. The ratings are from 1 to 5, where 1 denotes the worst and 5 denotes the best. They are also asked to write down their opinions on our synthesized fonts from any perspectives.\par
The results are shown in Table~\ref{tab:UserStudy}, where we employ three commonly-used metrics: precision, recall and accuracy to evaluate the performance of user study.
The precision of participants' choices is typically high but the recall is much lower ($<$45\% for both designers and non-designers).
It suggests that most participates can pick out a few machine-generated glyphs but there are also many synthesis results that cannot be distinguished from human-designed ones.
The designers perform better in finding synthesized fonts than non-designers because of their professional knowledge.
The average accuracy is around 60\%, which indicates that both the quality and style consistency of our synthesized fonts are comparable to human-designed vector fonts.
We can see that all participates give high ratings for the style consistency of our synthesized fonts.
Moreover, participants of non-designers are also satisfied with the quality of our synthesized glyphs. We notice that some professional font designers give slightly lower ratings on the font quality and suggest further improving the details of some machine-generated glyphs. Indeed, as discussed in Section 4.13, there still exist some limitations in the proposed method, which will be investigated in our future work.
\begin{table}
	\centering
	\caption{Results of the user study conducted to evaluate our synthsized fonts. "SC" denotes the style consistency of different glyphs in a single font. The ratings of "Quality" and "SC" are from 1 to 5.}
	\begin{tabular}{lccccc}
		\toprule
		User Group    & Precision   & Recall   & Accuracy & Quality$\uparrow$  & SC$\uparrow$ \\
		\midrule
		Designers     &  83.7\%   &  41.2\%  & 66.5\% & 3.4 & 4.5 \\
		Non-designers & 67.1\%   &  24.3\%  & 55.6\%  & 4.6 & 4.8 \\
		Average     &  77.0\%  & 34.4\%  & 62.1\% & 3.9 & 4.6 \\

		\bottomrule
	\end{tabular}
    \label{tab:UserStudy}
\end{table}

\subsection{Generating Fonts in Special Styles}
\begin{figure}[t!]
  \centering
  \includegraphics[width=\columnwidth]{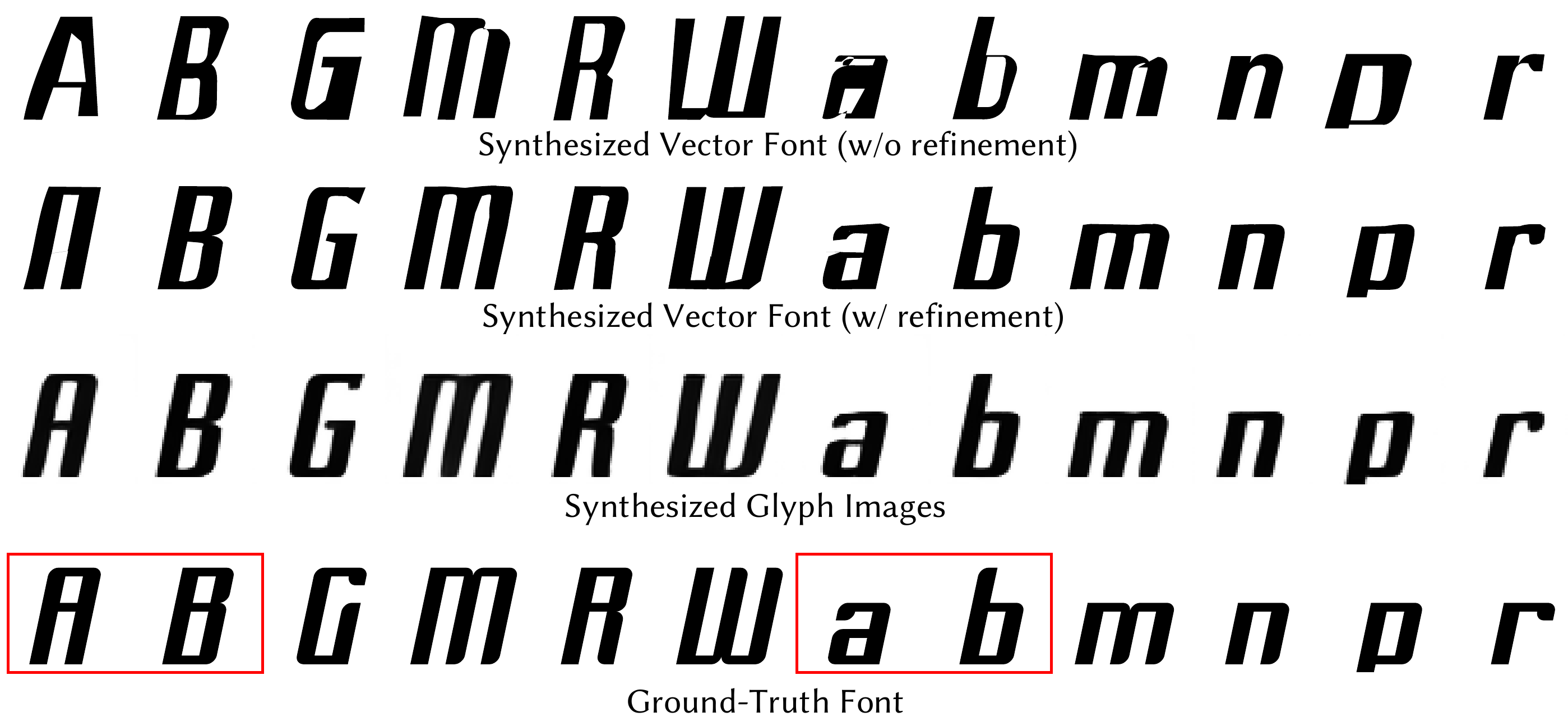}
  \caption{Generating fonts in special styles. The input reference glyphs are marked by red rectangles.}
  \label{fig:NovelFonts}
\end{figure}
Fig.~\ref{fig:NovelFonts} shows some synthesized vector glyphs of our model in a novel and special style.
The target font is selected from the testing dataset and possesses an unique style. 
Our synthesis results share the similar font style with the ground truth although there exist some differences in local details.
\subsection{Limitations}
\begin{figure}[t!]
  \centering
  \includegraphics[width=\columnwidth]{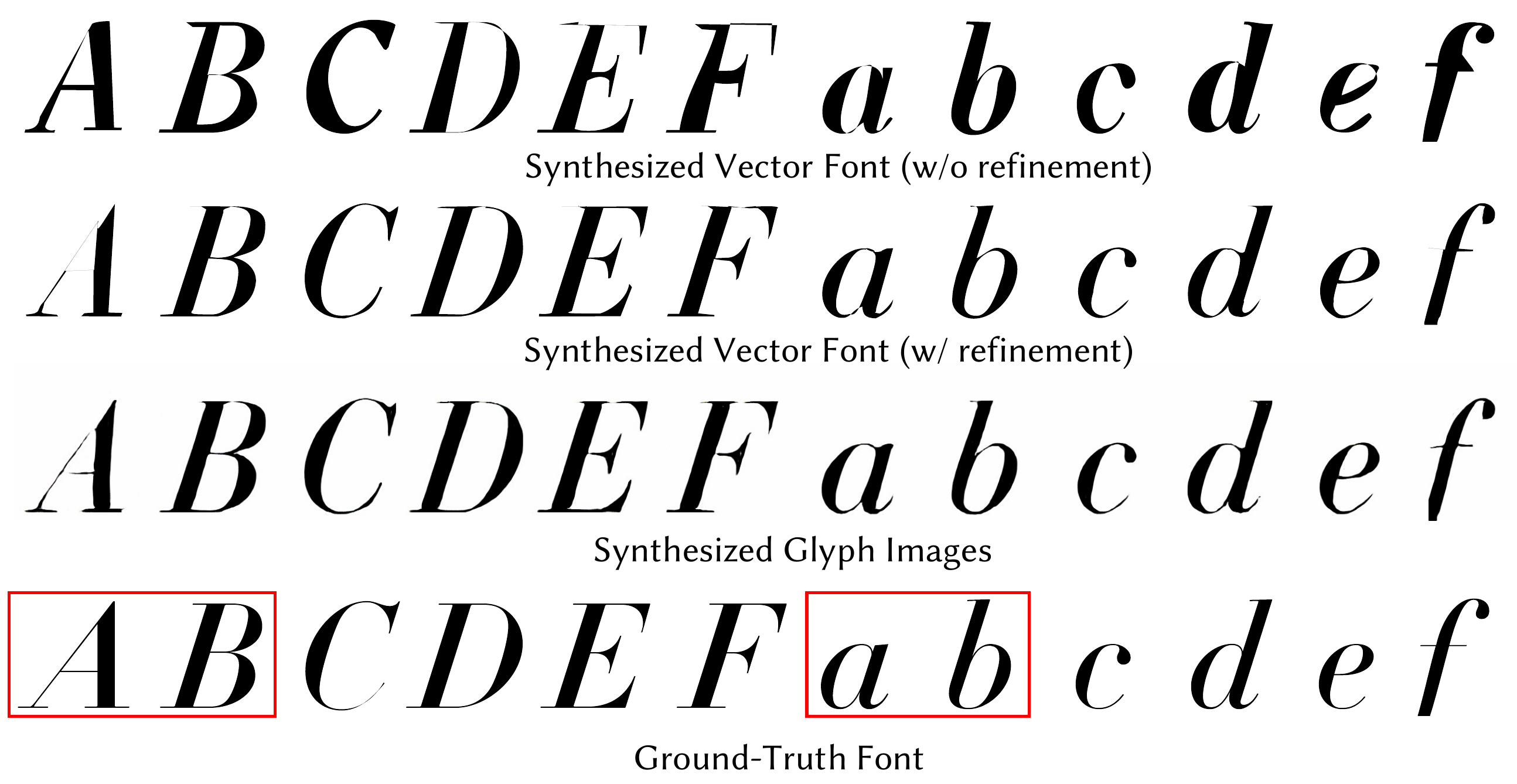}
  \caption{A typical failure case of our method when handling glyphs in extremely thin shapes. The input reference glyphs are marked by red rectangles.}
  \label{fig:Limitations}
\end{figure}
As shown in Fig.~\ref{fig:Limitations}, our model tends to generate broken outlines when handling glyphs that contain extremely thin strokes.
One reason is that the outlines of these synthesized glyph images are broken and the refinement process is guided by them.
Another reason is that argument points of outlines in this font are closely located, which confuses our model.
The problem can be resolved to some extent if we upgrade the size of our neural network model to accept and synthesize glyph images with higher resolution (e.g., 1024x1024).
Another limitation is that small curves tend to be overfitted to the counterparts in the generated images, which are usually not that smooth (e.g., the results of the first font in Fig.~\ref{fig:RefiningFromSimilarFonts}).
The problem is possible to be resolved by adding regularization in the loss functions during the refinement process to guarantee the smoothness of curves.
We leave these two issues as our future work.

\section{Conclusion}

In this paper, we presented a method that is able to synthesize high-quality vector fonts by exhaustively exploiting valuable and complementary information from the two different modalities (i.e., raster images and vector outlines) of fonts. To the best of our knowledge, this is the first work that is capable of automatically generating visually-pleasing vector glyphs whose quality and compactness are comparable to human-designed ones. Experiments conducted on a publicly-available dataset verified the effectiveness of our method and demonstrated that the proposed DeepVecFont obtains significantly better results compared to the state of the art both qualitatively and quantitatively. In the future, we are planing to upgrade our model by utilizing more powerful image and sequence synthesizing techniques to address the above-mentioned problems our current model has when handling fonts with very thin and artistic styles. Moreover, how to extend our method to more challenging font synthesis tasks for other writing systems (e.g., Chinese) is also an interesting research direction.

\begin{acks}
This work was supported by Beijing Nova Program of Science and Technology (Grant No.: Z191100001119077), Project 2020BD020 supported by PKU-Baidu Fund, National Language Committee of China (Grant No.: ZDI135-130), Center For Chinese Font Design and Research, Key Laboratory of Science, Technology and Standard in Press Industry (Key Laboratory of Intelligent Press Media Technology),  State Key Laboratory of Media Convergence Production Technology and Systems.
\end{acks}

\bibliographystyle{ACM-Reference-Format}
\bibliography{sample-bibliography}


\end{document}